\documentclass[11pt]{article}

\usepackage{url}
\usepackage{enumerate,mathrsfs,comment}
\usepackage[numbers]{natbib}
\usepackage[hypertexnames=false, colorlinks,linkcolor=red,anchorcolor=blue,citecolor=blue]{hyperref}
\usepackage{fullpage}

\usepackage[protrusion=false,expansion=true]{microtype}

\RequirePackage{amsmath}
\RequirePackage{amssymb}
\RequirePackage{amsthm}
\RequirePackage{bm}
\usepackage{multirow}
\usepackage{graphicx}
\usepackage{subfigure}
\usepackage{makecell}
\usepackage{booktabs}
\usepackage{array}
\usepackage{url}
\usepackage{algorithm}
\usepackage{algorithmic}
\usepackage{dsfont}

\let\tilde\widetilde

\newcommand{\cL}{\mathcal{L}}

\newcommand{\diag}{{\rm diag}}

\DeclareMathOperator{\E}{E}

\theoremstyle{plain}
\numberwithin{equation}{section}

\newtheorem{lemma}{\indent \bf Lemma}

\newtheoremstyle{mytheoremstyle} %
    {\topsep}                    %
    {\topsep}                    %
    {\normalfont}                   %
    {}                           %
    {\bfseries}                   %
    {.}                          %
    {.5em}                       %
    {}  %

\theoremstyle{mytheoremstyle}

\ifx\BlackBox\undefined
\newcommand{\BlackBox}{\rule{1.5ex}{1.5ex}}  %
\fi

\ifx\QED\undefined
\def\QED{~\rule[-1pt]{5pt}{5pt}\par\medskip}
\fi

\ifx\proof\undefined
\newenvironment{proof}{\par\noindent{\bf Proof\ }}{\hfill\BlackBox\\[2mm]}
\fi

\ifx\theorem\undefined
\newtheorem{theorem}{Theorem}
\fi
\ifx\example\undefined
\newtheorem{example}[theorem]{Example}
\fi
\ifx\property\undefined
\newtheorem{property}{Property}
\fi
\ifx\lemma\undefined
\newtheorem{lemma}[theorem]{Lemma}
\fi
\ifx\proposition\undefined
\newtheorem{proposition}[theorem]{Proposition}
\fi
\ifx\remark\undefined
\newtheorem{remark}[theorem]{Remark}
\fi
\ifx\corollary\undefined
\newtheorem{corollary}[theorem]{Corollary}
\fi
\ifx\definition\undefined
\newtheorem{definition}[theorem]{Definition}
\fi
\ifx\conjecture\undefined
\newtheorem{conjecture}[theorem]{Conjecture}
\fi
\ifx\fact\undefined
\newtheorem{fact}[theorem]{Fact}
\fi
\ifx\claim\undefined
\newtheorem{claim}[theorem]{Claim}
\fi
\ifx\assumption\undefined

\fi
\numberwithin{equation}{section}
\numberwithin{theorem}{section}

\providecommand{\keywords}[1]{\textbf{\textit{Keywords:}} #1}

\begin{document}

\title{Auto-encoding graph-valued data with applications to brain connectomes} 

\author
{
    Meimei Liu\thanks{Virginia Tech, Blacksburg, VA, 24060, United Staties. E-mail: meimeiliu@vt.edu.},    $\quad$
    Zhengwu Zhang\thanks{The University of North Carolina at Chapel Hill, Chapel Hill, NC, 27599, United States. E-mail: zhengwu\_zhang@unc.edu.},   $\quad$
    David B. Dunson\thanks{Duke University,Durham, NC, 27705, United States. E-mail: dunson@duke.edu. We acknowledge the funding from NIH R01-MH118927 and the Alibaba funding support.} 
}

\date{}
\maketitle

\begin{abstract}
There has been a huge interest in studying human brain connectomes inferred from different imaging modalities and exploring their relationships with human traits, such as cognition. Brain connectomes are usually represented as networks, with nodes corresponding to different regions of interest (ROIs) and edges to connection strengths between ROIs. Due to the high-dimensionality and non-Euclidean nature of networks, it is challenging to depict their population distribution and relate them to human traits. Current approaches focus on summarizing the network using either pre-specified topological features or principal components analysis (PCA). In this paper, building on recent advances in deep learning, 
we develop a nonlinear latent factor model to characterize the population distribution of brain graphs and infer their relationships to human traits. 
We refer to our method as Graph AuTo-Encoding (GATE). We applied GATE to two large-scale brain imaging datasets, the Adolescent Brain Cognitive Development (ABCD) study and the Human Connectome Project (HCP) for adults, to study the structural brain connectome and its relationship with cognition. Numerical results demonstrate huge advantages of GATE over competitors in terms of prediction accuracy, statistical inference, and computing efficiency. We found that the structural connectome  has a stronger association with a wide range of human cognitive traits than was apparent using previous approaches. 
\end{abstract}

\keywords{Brain networks; Non-linear factor analysis; Graph CNN; Replicated networks; Variational Auto-encoder.}

\section{Introduction}
Understanding the brain connectome and how it relates to human traits and various clinical variables has drawn huge attention \cite{park2013structural, fornito2013graph,craddock2013imaging,Jones2013}.  This has motivated large neuroimaging studies with thousands of subjects, such as the UK Biobank (UKB) \cite{miller2016multimodal},  the Adolescent Brain Cognitive Development (ABCD) study \cite{casey2018adolescent}, and the Human Connectome Project (HCP) \cite{van2013wu}. Through these studies, there have been dramatic improvements in the ability to reconstruct brain connectomes thanks to advanced hardware \cite{glasser2016human}, novel image acquisition protocols \cite{glasser2016human,Tuch2004}, and new reconstruction algorithms \cite{Zhang2017HCP,Smith2012}. In this paper, we are particularly interested in diffusion magnetic resonance imaging (dMRI), which is a commonly used  technique that measures the movement of water molecules along major fiber bundles in  white matter (WM) fiber tracts, enabling reconstruction of individual-level microstructural brain networks delineating anatomical connections between brain regions. This paper aims at developing advanced analysis methods for the brain structural connectomes recovered from diffusion MRI data.

Let $A_i$ represent the structural connectivity recovered from subject $i$, with element $A_{i[uv]}$ measuring white matter connections between brain regions $u$ and $v$. Using $n$ individual brain networks, we are interested in (a) 
appropriately summarizing each individual brain network in a parsimonious manner, isolating unique features of the network without discarding valuable information, (b) inferring relationships between brain networks and human traits, and (c) characterizing variation across individuals in their network structure.

There are existing methods  relevant to these goals. 
Based on a latent space characterization,  \cite{durante2017nonparametric} proposed a random effects model to represent the population distribution of brain networks.
Their approach clusters individuals based on brain structure and allows inferences on group differences \cite{durante2018bayesian}. Disadvantages include the highly computationally intensive implementation and coarse characterization of individual differences based on clustering. 
There is also literature on PCA-style approaches.  One possibility is to simply stack the adjacency matrices $A_i$ for individuals $i=1,\ldots,n$ into a tensor, and then apply tensor PCA and its variants to get summary scores of networks \cite{zhang2019tensor,zhang2017common}. These scores are treated as brain network surrogates in subsequent analyses, e.g., relating brain networks to human traits \cite{zhang2019tensor}. Tensor PCA is relatively efficient computationally while providing a simple low-dimensional summary of an individual's brain structure, but it is linear, limiting the ability to represent brain networks parsimoniously. Other matrix-based approaches, including spatial independent component analysis (ICA), non-negative matrix factorization (NMF), and spatial sparse coding algorithms, are also widely used in the analysis of functional brain connectomes; see \cite{beckmann2005investigations}, \cite{xie2017decoding}, and references therein. An alternative is graph representations based on geometric deep learning. \cite{kipf2016semi}, \cite{inductive} proposed graph convolutional networks (GCN) that use structure information in learning a low-dimensional feature representation for each node in a graph. \cite{kawahara2017brainnetcnn}, \cite{ktena2018metric}  applied GCN to functional brain connectomes for classification and similarity ranking. \cite{zhao2019variational} proposed a variational autoencoder-based Gaussian mixture model for functional brain connectomes classification. 
Advanced graph embeddings for structural brain connectomes are still lacking.

A major motivation of this article is to develop a non-linear latent factor modeling approach to (1) provide a characterization of the population distribution of brain graphs and (2) output low-dimensional features that can be used to summarize an individual’s graph. Compared with the original high-dimensional adjacency metrics, the low-dimensional features of brain networks can further facilitate visualization, prediction, and inference on relationships between connectomes and human traits. 
With this motivation, we are particularly intrigued by deep neural networks for non-linear dimension reduction. Generative algorithms, such as Variational Auto-Encoders (VAEs) \cite{kingma2014auto,rezende2014stochastic}, have proven successful in representing images via low dimensional latent variables. 
VAEs model the population distribution of image data through a simple distribution for the latent variables combined with a complex non-linear mapping function. A key to the success of such methods is the use of convolutional operators to encode symmetries often present in images. However, structural brain networks have a fundamentally different geometric structure, and such methods cannot be employed directly. 

We develop a model-based variational {\bf{G}}raph {\bf{A}}u{\bf{t}}o-{\bf{E}}ncoder (GATE) for brain connectome analysis. GATE consists of two components. The first component is a generative model that specifies how the latent variables $z_i$ give rise to the observations $A_i$ through a non-linear mapping, parametrized by neural networks. The second component is an inference model that learns the inverse mapping from $A_i$ to $z_i$. Our main contributions can be summarized as follows.

First, GATE learns the embedding and the population distribution of brain connectomes simultaneously. This is achieved by: 1) a nonlinear latent factor model to obtain a low-dimensional representation $z_i$ of brain network $A_i$; and 2) a hierarchical generative model designed to learn the conditional distribution $p(A_i\mid z_i)$ so that one can accurately reconstruct the brain network from the latent embedding.  
We model each cell $A_{i[uv]}$ in $A_i$ using a latent space model \cite{hoff2002latent}, with the latent coordinates of the regions $u$ and $v$ varying as a nonlinear function of the  individual-specific features $z_i$. This step involves a novel graph convolutional network that relies on the intrinsic locality of the brain networks to propagate node-specific $k$-nearest  neighbor information. 

Second, we extend GATE to relate human phenotypes to brain structural connectivity, which we refer to as  {\bf{re}}gression with {\bf{GATE}} (reGATE). reGATE is a supervised embedding method that simultaneously learns the population distribution of brain networks, network embeddings, and a predictive model for human traits. Although there has been some work integrating regression models and VAEs, the focus has been on multi-stage approaches; e.g., see \cite{yoo2017variational} as an example. reGATE can generate from the population distribution of brain networks conditionally on the value of a human trait. This provides invaluable information about how traits and brain networks are associated while characterizing variation across individuals.
 We further draw inference on selected network summary measures of interest, such as network density and average path length, to understand how these properties are distributed depending on human traits.

We apply GATE and reGATE to brain connectomes from ABCD and HCP and find strong relationships between structural connectomes and cognition traits in both datasets.
ReGATE shows superior performance in predicting the relationship between cognition and brain connectomes, particularly when trained with data from large numbers of individuals.
For example, using more than five thousand brain scans in the ABCD study, 
reGATE improves prediction of cognitive traits by $30\%-40\%$ compared with existing competitors.
Through detailed inference based on reGATE, we show that individuals with high cognitive traits tend to have denser connections between hemispheres, higher overall network density, and lower average path length. Such network summary measures have higher variability across the children evaluated in ABCD compared with adults in HCP.

\section{Methods}
\subsection{Brain Imaging Datasets and Structural Connectome Extraction}\label{subsec:data}
We focus on two large datasets in this paper: the Adolescent Brain Cognitive Development (ABCD) dataset and the Human Connectome Project (HCP) dataset. 

{\bf ABCD dataset}: The ABCD study in the United States focuses on tracking brain development from childhood through adolescence to understand biological and environmental factors that can affect the brain's developmental trajectory. The research consortium consists of 21 research sites across the country and invited 11,878 9-10-year-old children to participate. Researchers track their biological and behavioral development through adolescence into young adulthood. The dataset can be downloaded from NIH Data Archive (NDA, \url{https://nda.nih.gov}). The imaging protocol is harmonized for three types of 3T scanners: Siemens Prisma, General Electric (GE) 750, and Philips. We downloaded the structural T1 MRI and diffusion MRI (dMRI) data for 5253 subjects from the ABCD 2.0 release in NDA. The structural T1 images were acquired with an isotropic resolution of 1 mm$^3$. The diffusion MRI images were obtained based on imaging parameters: 1.7 mm$^3$ resolution, four different b-values ($b= 500, 1000, 2000, 3000$) and 96 diffusion directions. There are 6 directions at $b= 500$, 15 directions at $b= 1000$, 15 directions at $b= 2000$, and 60 directions at $b= 3000$. Multiband factor 3 is used for dMRI acceleration. See \cite{casey2018adolescent} for more details about data acquisition and preprocessing of the ABCD data. 

{\bf HCP dataset}: The HCP aims at characterizing human brain connectivity in about $1,200$ healthy adults to enable detailed comparisons between brain circuits, behavior, and genetics at the level of individual subjects \cite{VanEssen20122222}. Customized scanners were used to produce high-quality and consistent data to measure brain connectivity. The data containing various traits and MRI data can be easily accessed through \url{https://db.humanconnectome.org/}.

To obtain structural connectomes, we used a state-of-the-art dMRI data preprocessing framework -- population-based structural connectome (PSC) mapping (\cite{zhang2018network}). PSC uses a reproducible probabilistic tractography algorithm (\cite{Girard2014266, maier2016tractography}) to generate whole-brain tractography. PSC borrows anatomical information from high-resolution T1 images to reduce bias in the reconstruction of tractography. We used the Desikan–Killiany atlas (\cite{Desikan2006968}) to define the brain regions of interest (ROIs) corresponding to the nodes in the structural connectivity network. The Desikan–Killiany parcellation has 68 cortical surface regions with 34 nodes in each hemisphere. For each pair of ROIs, we extracted the streamlines connecting them. In this process, several procedures were used to increase reproducibility: (1) each gray matter ROI is dilated to include a small portion of white matter region, (2) streamlines connecting multiple ROIs are cut into pieces so that we can extract the correct and complete pathway and (3) outlier streamlines are removed.  We use the number of fibers connecting each pair of ROIs to summarize connectivity in our analyses.

\begin{figure}[htb!]
 \centering
\includegraphics[scale=0.45]{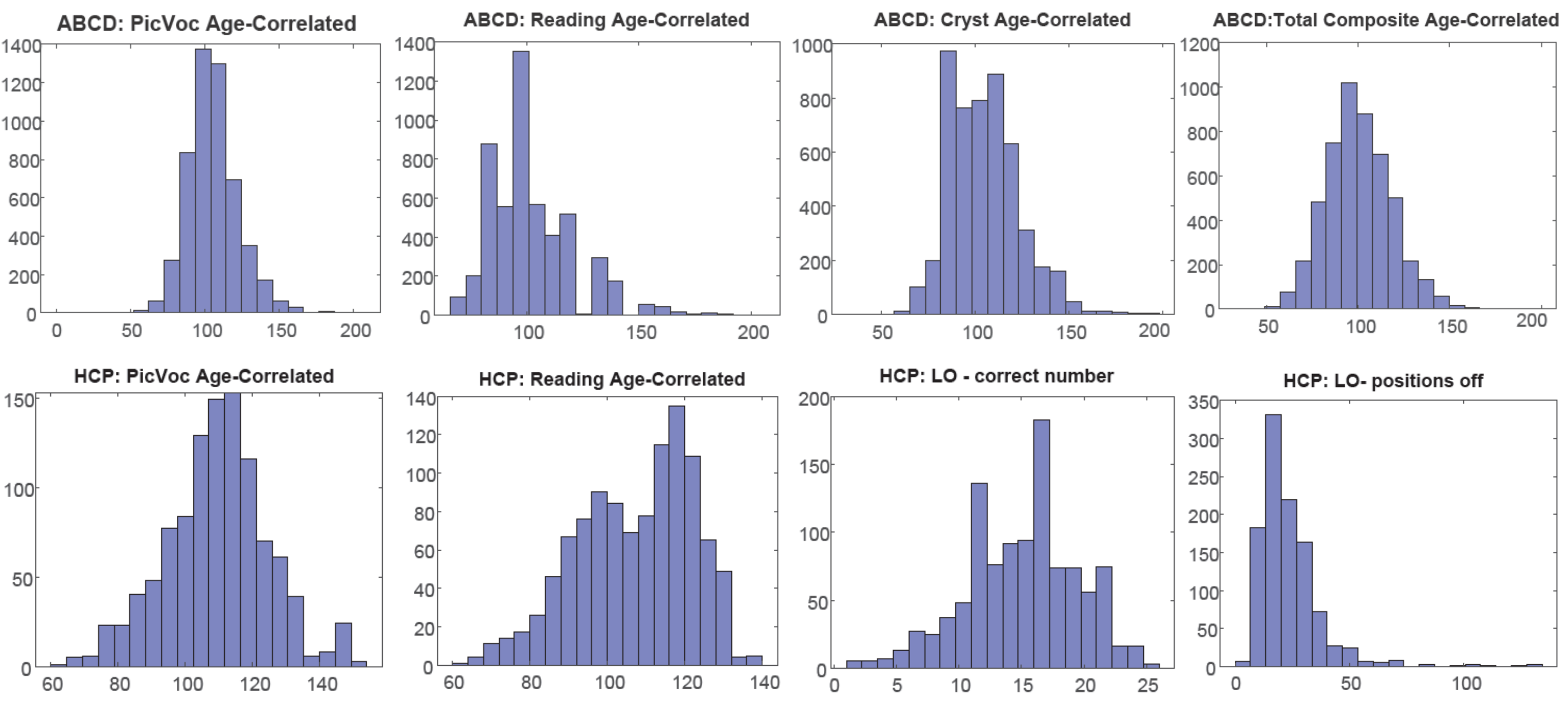}
  \caption{\small Histograms of cognitive traits. The first row: ABCD study with 5252 subjects involved; from left to right, the traits include picture vocabulary score, oral reading recognition test score, crystallized composite age-corrected standard score, cognition total composite score. 
  The second row: HCP study with 1065 subjects; from left to right: picture vocabulary test score,  oral reading recognition test score,  line orientation (LO) - total number correct, and line orientation (LO) - total positions off for all trials. 
   } \label{fig:real:abcd:traits_hist}
\end{figure}

For the ABCD dataset, we processed $5252$ subjects using PSC. We focus our analyses on four cognitive traits: (a) picture vocabulary score, (b) oral reading recognition test score, (c) 
crystallized composite age-corrected standard score, (d) cognition total composite score. The first row in Figure \ref{fig:real:abcd:traits_hist} demonstrates the distribution of the four traits. Similarly, for the HCP dataset, we preprocessed 1065 subjects using PSC. We focused on the cognitive traits: (a) picture vocabulary test score, (b) oral reading recognition test score, (c) line orientation - total number correct, and (d) line orientation - total positions off for all trials.

\subsection{The Graph Auto-Encoder Model}\label{sec:gate}
The brain connectome for individual $i$ is represented as a $V\times V$ symmetric adjacency matrix $A_i$, where $A_{i[uv]}$ 
is the count of the number of fibers connecting regions $u$ and $v$ in the $i$th individual's brain. Using the Desikan atlas, we choose $V=68$ ROIs. We let  
$$L(A_i) = (A_{i1},...,
A_{iV(V-1)/2}) \equiv (A_{i[21]}, A_{i[31]}, \cdots, A_{i[V1]}, A_{i[32]}, \dots, A_{i[V2]}, \dots, A_{i[V(V-1)]})$$ denote the lower triangular elements of matrix $A_i$. We let $y_i$ denote the value of a cognitive trait for individual $i$.
Our goal is to model the population distribution of the $A_i$'s and learn the relationship between cognitive traits $y_i$ and brain structural connectomes $A_i$. 

\subsubsection{Latent space model for brain connectomes}
Latent space models \cite{hoff2002latent} provide a probabilistic framework that assumes the edges in the networks are conditionally independent given their corresponding edge probabilities, with these probabilities defined as a function of pairwise distances between the nodes in a latent space. 
Borrowing the conditional independence idea, we first introduce a general latent space model for brain connectomes. We assume that the number of fibers connecting brain regions are conditionally independent Poisson variables, given individual- and edge-specific rates $\lambda_i = \{\lambda_{i1}, \cdots, \lambda_{i V(V-1)/2}\}^\top$, 
\begin{equation}\label{eq:model:general}
A_{i\ell}| \lambda_{i\ell} \sim  \; \textrm{Poisson}\big(\lambda_{i\ell}\big),
\end{equation}
independently for each pair $\ell = 1, \cdots, V(V-1)/2$ and $i=1, \cdots, n$. We assume $\log(\lambda_{i\ell})$ has the following factorization form: 
\begin{align}
 \log(\lambda_{i\ell}) = & \gamma_{\ell} + \psi_{ \ell}^{(i)} \label{eq:latent_decompose_1},\\
 \psi_{\ell}^{(i)} =& \sum_{r=1}^R \alpha_r X_{ur}^{(i)}X_{vr}^{(i)},\quad \mbox{for $\ell=[uv]$}, \label{eq:latent_decompose_2} \\
 \textrm{and} \; \; X_r^{(i)} = & (X_{1r}^{(i)}, \cdots, X_{Vr}^{(i)})^\top . \label{eq:latent_decompose_3}
\end{align} 
As shown in (\ref{eq:latent_decompose_1}), $\log(\lambda_{i\ell})$ is decomposed into two parts: a baseline parameter $\gamma_{\ell}$ controlling connection strength between the $\ell$th pair of brain regions, representing shared structure across individuals, and an individual deviation $\psi_{\ell}^{(i)}$. Taking into account symmetry constraints and excluding the diagonal elements, there are $V(V-1)/2$ unknown $\{\lambda_{il}\}$ for each subject,  leading to a daunting dimensionality problem. To reduce dimensionality,  \cite{durante2017nonparametric} proposed an SVD-type latent factorization, as shown in (\ref{eq:latent_decompose_2}), where $r=1,\ldots,R$ indexes the different latent dimensions, $\alpha_r>0$ is a weight on the importance of dimension $r$, and $X_{ur}^{(i)}$ is the $r$th latent factor specific to brain region $u$ and subject $i$. According to (\ref{eq:latent_decompose_2}-\ref{eq:latent_decompose_3}),
if $X_{ur}^{(i)}$ and $X_{vr}^{(i)}$ have the same sign and neither are close to zero, we have $X_{ur}^{(i)}X_{vr}^{(i)} > 0$ and there will be a positive increment on $\psi_\ell^{(i)}$ and hence on the expected number of fibers connecting regions $u$ and 
$v$ for subject $i$.
 
Model (\ref{eq:model:general})-(\ref{eq:latent_decompose_3}) can flexibly characterize variability across individuals in brain connectivity, while accommodating the complexity of network structures within each individual. However, we face challenges in learning the latent representations using existing latent space models: 1) Non-linearity: graph data are generally non-Euclidean with complicated structures. Designing a model to efficiently capture the non-linear structure is difficult. 2) Sparsity: brain regions are not fully connected, particularly structural brain networks. 3) speed: existing latent space approaches often rely on Markov chain Monte Carlo sampling, which is computationally intensive for high dimensional graphs. It is desirable to develop a fast non-linear factorization model. To address these challenges, we propose an autoencoder-based approach called Graph Autoencoder (GATE), from which we model the latent coordinates $X_u^{(i)}$ of brain regions as a non-linear function of a lower-dimensional vector $z_i$, which serves as a low-dimensional representation of $i$-th brain network $A_i$. 

\subsubsection{The Graph AutoEncoder (GATE) Model}\label{sec:GATE}
GATE relies on the variational autoencoder (VAE \cite{kingma2014auto}), which is a popular technique for non-linear dimension reduction. Denote $z_i\in \mathbb{R}^K$ as a low-dimensional latent representation of the individual brain connectome $A_i$. GATE consists of two components. The first component is a generative model that specifies how the latent variables $z_i$ give rise to the observations $A_i$ through a non-linear mapping, parametrized by neural networks. The second component is an inference model that learns the inverse mapping from $A_i$ to $z_i$. We frame our proposed Graph Autoencoder (GATE) method in the following context.  

 \subsubsection{Generative model} \label{sec:generative}
 For each subject $i$, we assume $A_{i\ell}$ for $\ell = 1,..., V(V-1)/2$ are conditionally independent given the latent representation $z_i \in \mathbb{R}^K$. 
Therefore, the likelihood of $L(A_i)$ is 
\begin{equation}\label{eq:gene_model_1}
p_\theta(L(A_i) = a_i | z_i) = \prod_{\ell=1}^{V(V-1)/2} p_\theta(A_{i\ell} = a_{i\ell} | z_i). 
\end{equation}
$p_\theta(A_{i\ell} | z_i)$ is a generative model for the weighted adjacency matrix $A_i$ given the 
latent $z_i$ with $z_i\sim p(z)$. We define $p(z) = N(0, I_K)$, representing all connectomes in the same Gaussian latent space. 

We learn the mapping from the Gaussian latent space to the complex observation distribution in (\ref{eq:gene_model_1}) by a hierarchical model equipped with parameters $\theta$. Specifically, we assume the observations $A_{i\ell}$ arise from the following generative process:
\begin{align}
z_i \sim &  \; N(0, I_K),  \nonumber\\
A_{i\ell}| z_i \sim & \; \textrm{Poisson}\big(\lambda_{i\ell}(z_i)\big), \label{eq:gene_model_0}
\end{align} 
where the Poisson rate parameter $\lambda_{i\ell}(z_i)$ is modeled as a nonlinear function of $z_i$ according to: 
\begin{align}
\lambda_{i\ell}(z_i) = & \exp(\gamma_{\ell} + \psi_{ \ell}(z_i)) \label{eq:decompose_1},\\
 \psi_{\ell}(z_i) =& \sum_{r=1}^R \alpha_r X_{ur}(z_i)X_{vr}(z_i),\quad \mbox{for $\ell=[uv]$}, \label{eq:decompose_2} \\
 X_r(z_i) = & (X_{1r}(z_i), \cdots, X_{Vr}(z_i))^\top = g_r(z_i), \label{eq:decompose_3}
\end{align}  
where $g_r(\cdot) : \mathbb{R}^K \to \mathbb{R}^{V}$ is a nonlinear mapping from $z_i$ to the $r$th latent factor of the brain regions $X_r$, parameterized by deep neural networks with parameters $\theta$ for $r=1, \cdots, R$.

Denote $\bold{X}(z_i) = (X_1(z_i), \dots, X_R(z_i)) \in \mathbb{R}^{V\times R}$. The $u$-th row $\big(X_{u1}(z_i),\dots, X_{uR}(z_i)\big)$ represents the latent features of brain region $u\in \mathcal{V}$ for individual $i$. A relatively large positive value for the cross product between the $u$-th and $v$-th rows implies a relatively high connection strength between these brain regions. The nonlinear mapping $\{g_r(\cdot)\}$ ($r=1,\dots, R$) characterizes the latent embedding of brain regions that is determined via the local collaborative patterns among brain regions. 
 
To take into account the intrinsic locality of structural brain networks, we propose a novel graph convolutional network (GCN) to learn each region's representation by propagating node-specific $k$-nearest neighbor information. Information in nodes that are closer to each other will be pooled together in GCN. The intrinsic locality refers to the relative distance between brain regions measured through the length of white matter fiber tracts connecting them. We extract this information from brain imaging tractography and store it in a matrix $B \in \mathbb{R}^{V\times V}$, where $B_{uv}$ is the averaged length of fiber tracts between region $u$ and $v$, $B_{uv} = B_{vu}$, $B_{uu}=0$, and we set $B_{uv}=\infty$ if there are no fibers between them. For each region $u$, we define its $k$-nearest neighbors ($k$-NN$(u)$) as the $k$ ROIs closest to $u$ according to our notion of distance, and denote the region itself as its $0$-NN. If a region $u$ has less than $k$ direct neighbors, we will include all the regions $v$ satisfying $B_{uv} \in (0,\infty)$ as its neighbors. In practice, we choose the average degree of nodes as the number of neighbors for simplicity since it measures the average number of collaborations among nodes.

To learn the $r$-latent coordinate $g_r(z_i)$ for subject $i$, the key idea is to consider each column $X_r(z_i)$ as an ``image'' with each region as an irregular pixel; we have $R$ such ``images'' for each individual. Convolutional neural networks (CNN) are highly effective architectures in image and audio recognition tasks \cite{krizhevsky2012imagenet, sermanet2012convolutional, hinton2012deep}, thanks to their ability to exploit the local translational invariance structures over their domain. Considering the unique features of the brain connectome networks, we generalize the CNN and define appropriate graph convolutions to learn the nonlinear mapping $\{g_r(\cdot)\}$ via exploiting the local collaborative pattern among brain regions. In particular, we define an $M$-layer GCN as follows:
\begin{align}
X_r^{(i,1)} = & h_1 (W^{(r,1)}z_i + b_1), \label{eq:init}\\
X_r^{(i,m)} = & h_m (W^{(r,m)} X_r^{(i,m-1)} + b_m) \quad \textrm{for}\; 2\leq m \leq M, \label{eq:layers}
\end{align}
where $X_r^{(i,m)}$ denotes the output of the $m$-th layer of the convolutional neural network, $h_m(\cdot)$ is an activation function for the $m$th layer, and $W^{(r,m)}$ is a weight matrix characterizing the convolutional operator at this layer. We denote the parameters $b_m$, $W^{(r,m)}$, together with $\gamma_\ell$, $\alpha_r$ in (\ref{eq:decompose_1}-\ref{eq:decompose_2}) ($m= 1, \dots, M, r=1,\dots, R, \ell =1, \dots, V(V-1)/2)$) as the model parameter $\theta$. 
The activation functions $\{h_m(\cdot)\}$ can be chosen from the following candidates based on performance: (1) rectified linear unit (ReLU) function, which is widely used \cite{goodfellow2016deep} in deep neural networks, with the definition as $ReLU (x) = \max (0, x)$, where the max operation is applied element-wise; (2) Sigmoid function defined as $h_m(x) = \frac{1}{1+e^{-x}} \in (0,1)$; (3) linear or identity function $h_m(x) = ax$ with $a\neq 0$. 

For $m = 1$, $ W^{(r,1)} \in \mathbb{R}^{V\times K}$ maps the latent representation $z_i\in \mathbb{R}^{K}$ to the latent space $X_r^{(i,1)}\in \mathbb{R}^{V\times 1}$; for $m\geq 2$, $W^{(r,m)}$ is a $V\times V$ weight matrix with the $u$-th row $w_{u\cdot}^{(r,m)}$ satisfying $w_{uv}^{(r,m)} >0$ if $v=u$ or $v\in k_r$-NN$(u)$, and $=0$ otherwise. 
(\ref{eq:layers}) implies that the embedding feature of each region at the $m$-th layer is determined by the weighted sum of itself and its nearest neighbor regions at the $(m-1)$-th layer, and the related weights aim to characterize the region-specific local connectivity. For $r=1,\cdots, R$, we can choose different values of $k$ to define its $k_r$-NN to fully explore the possible collaboration pattern among brain regions. 

\begin{figure}[h!]
\centering
\includegraphics[width=\textwidth]{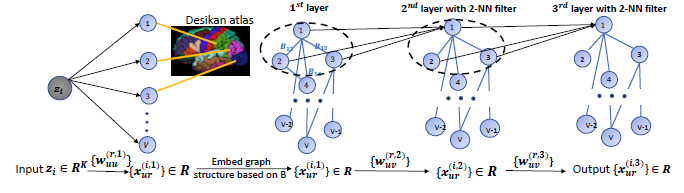}
\vspace{-10pt}
\caption{\small Illustrative example of three-layer GCN architecture with 2-NN filters. For example, to learn the $r$-th latent coordinate for node $1$, the $2$-NN is node $2$ and node $3$. After input $z_i$, the first layer embedding is  $x_{1r}^{(i,1)}= h_1(w_{11}^{(r,1)}z_i)$, then the second layer embedding is  $x_{1r}^{(i,2)}=h_2(w_{11}^{(r,2)}x_{1r}^{(i,1)}+w_{12}^{(r,2)}x_{2r}^{(i,1)} + w_{13}^{(r,2)}x_{3r}^{(i,1)})$, and the third layer embedding is $x_{1r}^{(i,3)}=h_3(w_{11}^{(r,3)}x_{1r}^{(i,2)}+w_{12}^{(r,3)}x_{2r}^{(i,2)} + w_{13}^{(r,3)}x_{3r}^{(i,2)})$, with the output $x_{1r}^{(i,3)}$ as the $r$-th latent coordinate for node $1$. }
\label{fig:knn:eg} 
\end{figure}

Figure \ref{fig:knn:eg} shows how a three-layer GCN learns $X_r(z_i)$ via a $2$-NN GCN. First, we initialize the latent feature for each region as $x_{ur}^{(i,1)}$ based on (\ref{eq:init}). 
Then, we construct a ``graph'' based on the fiber length in $B$: each region is assigned to connect with $2$ nearest neighbors at most according to the fiber length to other brain regions. 
This information is reflected in $W^{(i,m)}$, whose rows contain at most three non-zero elements (one at the diagonal and two off the diagonal). 
Next, we update the latent feature of each region in the next layer based on a sum of reweighted features from its $2$ nearest neighbors and itself. 

We collect all the parameters $(\gamma_\ell, \alpha_r, b_m, W^{(r,m)})$ for $\ell= 1, \dots, V(V-1)/2, r=1,\dots, R, m=1,\dots, M$ as $\theta$. In the following Section \ref{subsec:gate_learning}, we show how to use variational inference to learn $\theta$. 

\subsubsection{Variational Inference and GATE Learning}\label{subsec:gate_learning}
To train and evaluate the deep generative model in (\ref{eq:gene_model_0}), we need to estimate $\theta$, the parameters characterizing the mapping from $z_i$ to $A_{il}$ and $p_\theta(z_i |A_i)$, the posterior distribution of the latent variable. By applying Bayes' rule, we have the posterior as 
$$
p_\theta(z_i |A_i) = \frac{p_\theta (A_i |z_i) p(z_i)}{p_\theta(A_i)}.
$$
Since the likelihood function $p_\theta (A_i |z_i)$ is parameterized via the neural network with non-linear transformations, both the marginal distribution $p_\theta(A_i)$ and the posterior probability distribution $p_\theta (z_i | A_i)$ are intractable. 
Hence, we resort to variational inference (VI) (\cite{jordan1999introduction}, \cite{hoffman2013stochastic}), a widely-used tool for approximating intractable posterior distributions. 
VI seeks a simple distribution $q_\phi(z_i|A_i)$ parameterized by $\phi$ from a variational family, e.g., a Gaussian distribution family, that best approximates $p_\theta(z_i |A_i)$. We call such $q_\phi(z_i|A_i)$ as the probabilistic encoder, which maps the input $A_i$ to a low dimensional latent representation $z_i$. The approximated posterior $q_\phi(z_i | A_i)$ should be close to $p_\theta (z_i|A_i)$. We use Kullback-Leibler (KL) divergence to quantify the separation between these two distributions, which is defined as $D_{KL} (Q|| P)= \E_{z\sim Q} \log \frac{Q(z)}{P(z)}$, measuring how much information is lost if the distribution $Q$ is used to represent $P$. We choose 
\begin{equation}\label{eq:q}
q_\phi(z_i|A_i) \sim N(\mu_\phi(A_i), \textrm{diag}\{\sigma^2_\phi (A_i)\}), 
\end{equation}
i.e., a fully factorized (diagonal covariance) Gaussian distribution, to facilitate computation. We design deep neural networks to learn the parameters in $\mu_\phi$ and $\sigma_\phi^2$, and denote the  parameters involved in deep neural networks as $\phi$. The details are in Supplementary  \ref{subsec:appendix:q}.

Our objective is to maximize the observed data log-likelihood $\log p_\theta(A_i)$, and also minimize the difference between the true posterior $p_\theta(z_i|A_i)$ and approximated posterior distribution $q_\phi(z_i|A_i)$. We express the above objective as 
\begin{align}
& \log p_\theta(A_i) - D_{KL} (q_\phi(z_i|A_i) || p_\theta(z_i|A_i)) \nonumber \\
=& \E_{q_\phi(z_i|A_i)} [\log p_\theta(A_i | z_i)] - D_{KL}(q_\phi(z_i|A_i) || p_\theta(z_i)) :=- \mathcal{L}(A_i; \theta, \phi),\label{eq:elbo:0}
\end{align}
where detailed calculation of (\ref{eq:elbo:0}) can be found in Supplementary \ref{subsec:apendix:elbo1}.  
Since $D_{KL} (q_\phi(z_i|A_i) || p_\theta(z_i|A_i))$  is nonnegative, $-\mathcal{L}(A_i; \theta, \phi)$ can be viewed as a lower bound on the marginal log-likelihood, referred to as the evidence lower bound (ELBO), which is a function of both $\theta$ and $\phi$. 
Therefore, the training objective is minimizing the negative of the ELBO, i.e., minimizing 
 \begin{equation}\label{eq:elbo:gate}
 \mathcal{L}(A_i; \theta, \phi) = - \E_{q_\phi(z_i|A_i)} [\log p_\theta(A_i | z_i)] + D_{KL}(q_\phi(z_i|A_i) || p(z_i)). 
 \end{equation}
 $\mathcal{L}(A_i; \theta, \phi)$ consists of two parts: the first term is the reconstruction error, measuring how well the model can reconstruct $A_i$; while the second term, defined as the KL divergence of the approximate posterior from the prior, is a regularizer that pushes $q_\phi(z_i|A_i)$ to be as close as possible to its prior $N(0,I_K)$. 

\begin{figure}[htb]
\centering
\includegraphics[width=0.7\textwidth]{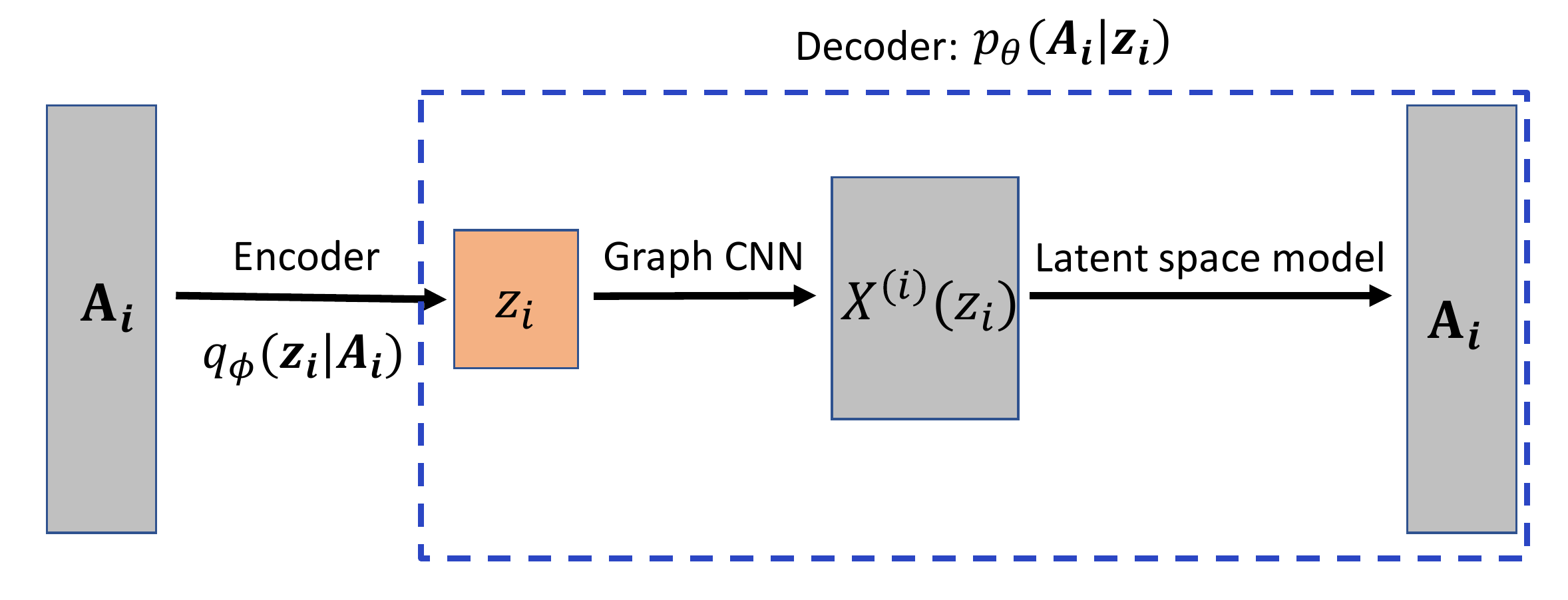}
\caption{A taxonomy of GATE. The encoder step is based on equation (12) to map $A_i$ to $z_i$. The decoder step is based on equation (6)-(9) to map $z_i$ to $A_i$ through a latent space model with node embedding $X^{(i)}(z_i)$ learned from a Graph CNN.}\label{fig:vae_flow_chart}
\end{figure}
In practice, the expectation in the ELBO (\ref{eq:elbo:gate}) is intractable. To address this, we employ Monte Carlo variational inference \cite{kingma2014auto} by approximating the troublesome expectation with samples of the latent variables from the variational distribution $z_i\sim q_\phi(z_i | A_i)$. Particularly, we form the Monte Carlo estimates of the expectation as 
$$
\E_{q_\phi(z_i|A_i)} [\log p_\theta(A_i | z_i)] \simeq \frac{1}{L} \sum_{\ell=1}^L \log p_\theta(A_i | z_i^{\ell}),
$$
where $z_i^{\ell}$ is sampled with the reparametrization trick: sampling $\varepsilon_i^{\ell}\sim N(0,I_K)$ and reparametrizing $z_i^{\ell} = \mu_\phi(A_i) + \varepsilon_i^{\ell} \odot \Sigma_\phi(A_i)$, where $\Sigma_\phi(A_i) = \textrm{diag}\{\sigma_\phi^2(A_i)\}$. A simple calculation shows that the KL-divergence $D_{KL}(q_\phi(z_i|A_i) || p(z_i))= \frac{1}{2}\sum_{k=1}^K \big(\mu_k^2 + \sigma_k^2 - 1 - \log (\sigma_k^2)\big)$, where $\mu_k$ and $\sigma_k$ are the $k$-th element of $\mu_\phi(A_i)$ and $\Sigma_\phi(A_i)$ respectively. Therefore, the ELBO in (\ref{eq:elbo:gate}) can be approximated as 
$$
\cL(A_i; \theta, \phi) \simeq \tilde{\cL}(A_i; \theta, \phi) = - \frac{1}{L} \sum_{\ell=1}^L \log p_\theta(A_i | z_i^{\ell}) + \frac{1}{2}\sum_{k=1}^K \big(\mu_k^2 + \sigma_k^2 - 1 - \log (\sigma_k^2)\big),
$$
which is differentiable with respect to $\theta$ and $\phi$. Then, given $n$ observed networks, we can construct an estimator of the ELBO of the full dataset, based on the minibatches $\frac{n}{m}\sum_{i=1}^m \tilde{\cL}(A_{(i)}; \theta, \phi)$, where $\{A_{(i)}\}_{i=1}^m$ is a randomly drawn sample of size $m$ from the full observed data with sample size $n$. 
Viewing $\frac{n}{m}\sum_{i=1}^m \tilde{\cL}(A_{(i)}; \theta, \phi)$  
as the objective, we implement a stochastic variational Bayesian algorithm to optimize $\theta$ and $\phi$, respectively. Figure \ref{fig:vae_flow_chart} shows the graphical diagram of the GATE approach; Algorithm \ref{table:GATE_alg} summarizes the GATE training procedure. 
Once the GATE model is learned, we can do the following: (1) have a low-dimensional representation for each individual network; (2) generate brain networks to learn the population distribution of  brain connectomes and features of these connectomes.

\begin{table}[htb]
\begin{center}
  \begin{tabular}{ l  c  r }
    \hline \hline
   {\bf{Input}}: $\{A_i\}_{i=1}^n$, $\{z_i\}_{i=1}^n$, geometric matrix $B$, latent space dimension $R$.  \\ \hline
  Randomly initialize $\theta, \phi$\\
  {\bf{while}} not converged {\bf{do}}\\
  \quad \quad Sample a batch of $\{A_i\}$ with mini-batch size $m$, denote as $\mathcal{A}_{m}$. \\
  \quad \quad \quad \quad {\bf{for all}} $A_i\in \mathcal{A}_m$ {\bf{do}} \\
   \quad \quad \quad \quad \quad \quad Sample $\varepsilon_i \sim N(0, I_R)$, and compute $z_i = \mu_\phi(A_i) + \varepsilon_i \odot \Sigma_\phi(A_i)$. \\
    \quad \quad \quad \quad \quad \quad Compute the gradients $\nabla_\theta \tilde{\cL}(A_{i}; \theta, \phi)$ and $\nabla_\phi \tilde{\cL}(A_{i}; \theta, \phi)$ with $z_i$. \\
    \quad \quad \quad \quad Average the gradients across the batch.\\
  \quad \quad Update $\theta$, $\phi$ using gradients of $\theta$, $\phi$. \\
  {\bf{Return}} $\theta, \phi$. \\ \hline   
   \hline 
  \end{tabular}
  \caption{\small Algorithm 1: Training GATE model using gradients.}\label{table:GATE_alg}
\end{center} 
\end{table}

\subsection{Regression with GATE and Inference}\label{sec:regate}
\begin{figure}[htb]
\centering
\includegraphics[width=0.7\textwidth]{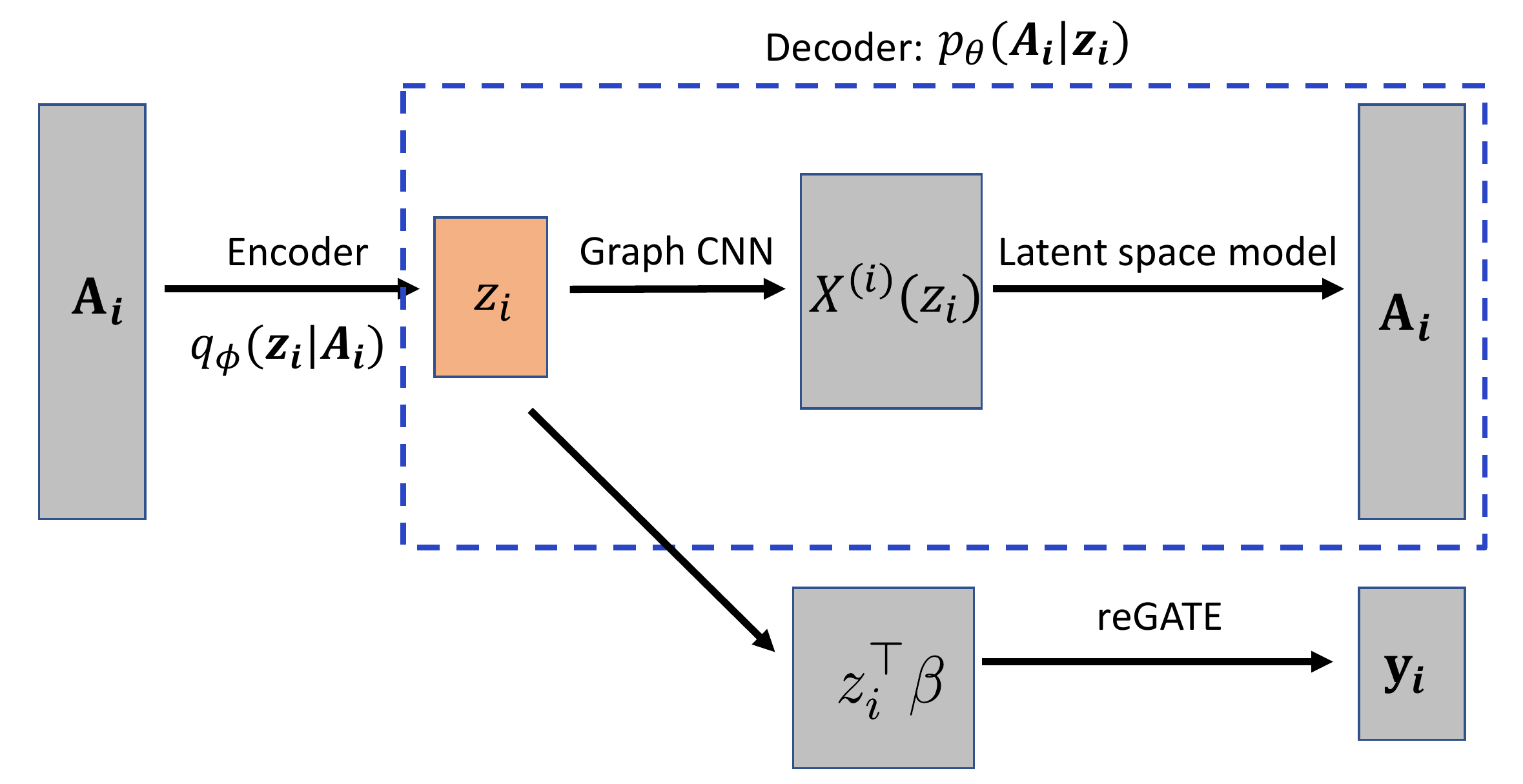}
\caption{reGATE to predict human traits. The  encoder step is based on equation (12) to map $A_i$ to $z_i$.The decoder step maps $z_i$ to $A_i$ based on equation (6)-(9); and a regression of $y_i$ with respect to $z_i$. } \label{fig:full_flow_chart}
\end{figure}

{\bf Relating brain connectomes with traits}. In addition to finding low-dimensional representations of brain structure networks, we are also interested in inferring the relationship between brain networks and human traits, such as cognition. With this goal in mind, we develop a supervised version of GATE, referred to as 
regression GATE (reGATE).
Let $y_i$ be a trait of the $i$-th subject. We first express the joint log likelihood of $(A_i, y_i)$ as 
\begin{align}
\log p_\theta(A_i, y_i) 
  = & -  \mathcal{L}(A_i, y_i; \theta, \phi) + D_{KL}(q_\phi(z_i|A_i) || p_\theta(z_i|y_i, A_i)), \label{eq:reg_elbo:1} 
\end{align} 
where
$$-  \mathcal{L}(A_i, y_i; \theta, \phi)= \E_{q_\phi(z_i|A_i)}\log p_\theta(y_i|z_i) + \E_{q_\phi(z_i|A_i)} \log p_\theta(A_i | z_i)  - D_{KL}(q_\phi(z_i|A_i)||p_\theta(z_i))$$
is called the ELBO of $\log p_\theta(A_i, y_i) $. We assume the human trait $y_i$ and the brain connectivity $A_i$ are conditionally independent given the latent representation $z_i$ for the $i$-th subject, and show the derivation of (\ref{eq:reg_elbo:1}) in Supplementary \ref{subsec:dev:elbo:regression}. In (\ref{eq:reg_elbo:1}), 
we divide the log-likelihood of $(A_i, y_i)$ into two parts: the ELBO denoted as $-\mathcal{L}(A_i, y_i; \theta, \phi)$, and the non-negative KL-divergence between $q_\phi(z_i|A_i)$ and $p_\theta(z_i|y_i, A_i)$.  
Different from the unsupervised ELBO in (\ref{eq:elbo:0}), (\ref{eq:reg_elbo:1}) can be considered as a supervised ELBO with an extra term $p_\theta (y_i |z_i)$ that essentially formulates a regression of $y_i$ with respect to $z_i$. 
Here we consider $y_i$ as a continuous random variable, and set $p_\theta (y_i |z_i)$ as a univariate Gaussian, i.e., $p_\theta(y_i | z_i)\sim N(z_i^\top \beta + b, \sigma^2)$, where $\beta, \sigma^2 \in \theta$ are parameters to be learned. Figure \ref{fig:full_flow_chart} shows the flowchart of the reGATE architecture. 

Similarly to Section \ref{subsec:gate_learning}, we form the Monte Carlo estimate of $\mathcal{L}(A_i, y_i; \theta, \phi)$ 
and estimate $\theta, \phi,\beta, b$ following the stochastic variational Bayesian Algorithm \ref{table:GATE_alg} by replacing $\cL(A_i; \theta, \phi)$ with $\cL(A_i,y_i; \theta, \phi)$. We show the detailed sampling steps in Supplementary  \ref{subsec:app:elbo:regression}. For the trained reGATE model, we have: (1) a low-dimensional representation for each individual network; (2) human trait prediction for each individual network; (3) ability to generate brain networks for inference on how features of the networks vary across individuals and with traits.

{\bf Conditional generative model}: We are interested in inferring how brain networks vary across levels of a trait.
 For example, if $y_i$ measures a person's memory ability, we would like to study differences in the distribution of brain networks 
 between people with good and bad memory skills.  
To address questions of this type, we generate samples from the posterior distribution of $A_i$ given particular $y_i$ values using Gibbs sampling: 
 sample $z_{i}$ from $p_\theta(z_i | y_i )$, then sample $A_{i}$ from $p_\theta(A_i|z_i)$. The conditional $p_\theta(A_i|z_i)$ is learned while implementing reGATE. The posterior distribution 
 of $z_i$ given $y_i$ can be expressed as $p_\theta(z_i | y_i ) \propto \; p_\theta(y_i | z_i) p_\theta(z_i)$, where $p_\theta(z_i) \sim  N(0, I_K)$ and $p_\theta(y_i |z_i) \sim  N(z_i^\top \beta + b, \sigma^2)$. Therefore, we have $p_\theta (z_i | y_i) \sim N\big(\mu_z(y_i), \Sigma_z(y_i)\big)$, where 
\begin{equation}\label{eq:z_to_y:0}
\mu_z(y_i) = (I_K + \beta \beta^\top /\sigma^2)^{-1} \beta (y_i-b) /\sigma^2, \quad\quad \textrm{and} \quad \Sigma_z(y_i)=\big(I_K + \beta \beta^\top /\sigma^2\big)^{-1};
\end{equation}
the derivation of 
 (\ref{eq:z_to_y:0}) is in Supplementary \ref{subsec:post:z:y}. 

The latent representation $z_i$ is unidentifiable in VAE since the log-likelihood and ELBO are rotationally invariant for $z_i$. For example, letting $\tilde{z}_i = U^\top z_i$, then $P_{U,\theta}(A_i) = P_\theta(A_i)$ and
\[D_{KL}(q_{U, \phi}(z_i|A_i) || p_{U,\theta}(z_i |A_i)) = D_{KL}(q_\phi(z_i|A_i) || p_\theta(z_i|A_i)),\]
where $U$ is an orthogonal matrix, $q_{U, \phi}(\cdot)$ and $p_{U,\theta}(\cdot)$ are defined by replacing $z_i$ with $\tilde{z}_i$ in $q_{\phi}(\cdot)$ and $p_{\theta}(\cdot)$. Rotational invariance can be solved by post-processing to rotationally align the $z_i$'s. However, this would only be necessary if one is attempting to compare $z_i$s from different datasets or analyses of a given dataset. Within an analysis, the main focus is on inference on the relative values of $z_i$s, and these relative values are well defined. In addition, when the focus is on relating brain structure to human traits or in predicting traits based on brain structure or vice versa, the non-identifiability issue does not present a problem.

\section{Simulation Study} \label{sec:sim}
\begin{figure}[h!]
\centering
\includegraphics[width=5in]{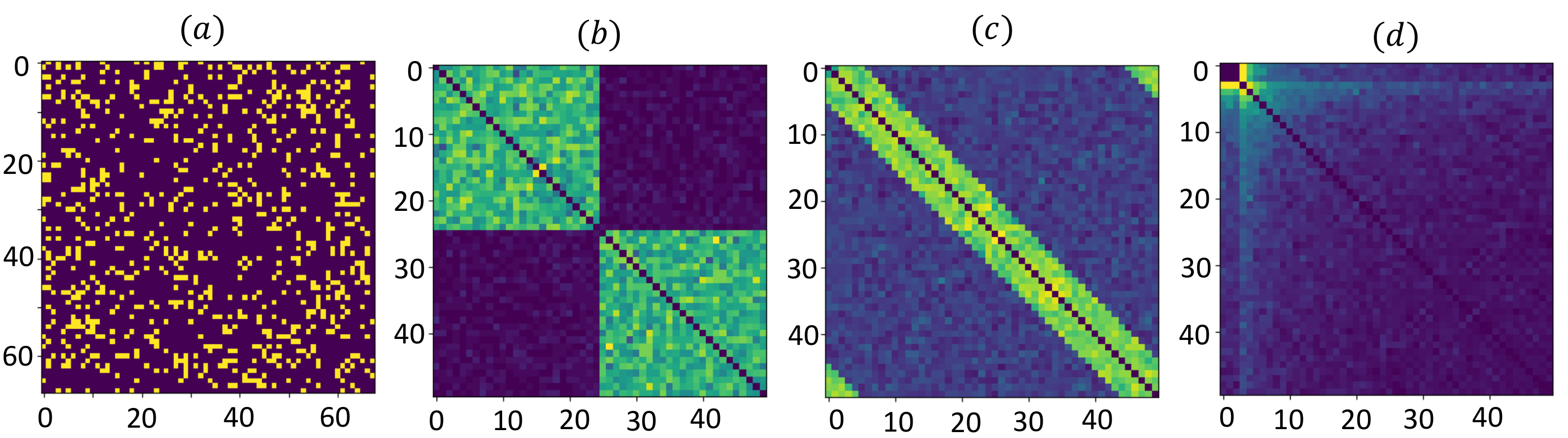}
\caption{The edge probability vectors (rearranged in matrix form) from four network structures: $(a)$ sparse graph, $(b)$ community structure, $(c)$ small world, $(d)$ scale free. } \label{fig:sim:original}
\end{figure}

We conduct a simulation study to evaluate the performance of GATE and reGATE on a broad application in graph-value data. 
We simulate different types of random graphs using the Python package {\it{NetworkX}}. Particularly, we consider four network structures: sparse networks according to the model in \cite{johnson1977efficient}, community structures under the model of \cite{nowicki2001estimation}, small-worldness from the model in \cite{watts1998collective}, and scale-free property from the model in \cite{barabasi1999emergence}. We simulate $100$ networks with $V=68$ nodes for each type by sampling their edges from conditional independent Bernoulli random variables given their corresponding structure-specific edge probability. Each structure-specific edge probability vector is carefully constructed to assign a high probability to a subset of network configurations characterized by a specific property. Figure \ref{fig:sim:original} displays some example networks we generated with the four different network structures. 
 
\begin{figure}[h]
\centering
\includegraphics[width=5in]{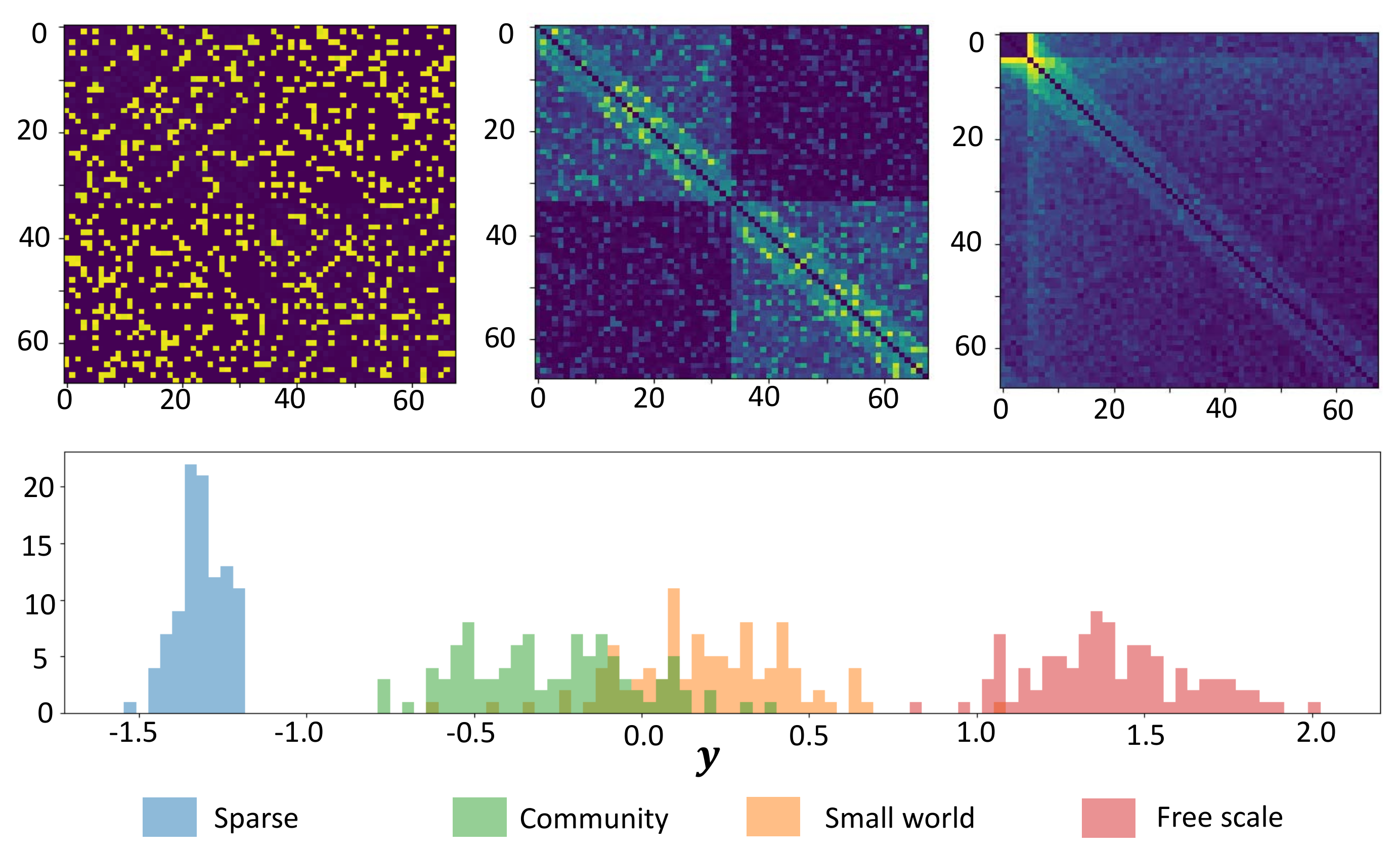}
\caption{\small First row: generated networks conditional on the specific $y_i$ using reGATE, corresponding to $y_i=-1.5, -0.1, 2$ respectively from left to right. Second row: histogram of $y_i$ with respect to the network structure; the x-axis is the value of $y_i$, the y-axis is the frequency that belongs to a specific structure in the training data. } \label{fig:sim:generate}
\end{figure}

 We first generate $y_i$ according to 
 $y_i = \alpha^\top A_i \alpha + \epsilon_i$, where $\alpha= (\underbrace{1,\cdots, 1}_{17}, 0, \cdots, 0)^\top \in \mathbb{R}^{68}$, and $\epsilon_i \sim N(0,1)$. We then standardize $y_i$, so that it ranges from $-1.5$ to $2.0$. These settings aim to generate separable $y_i$'s according to the topological structures of $A_i$. The histograms in Figure \ref{fig:sim:generate} clearly demonstrate how $y_i$ varies for different network structures. Our goals in this simulation study include (1) learning the latent representation under both GATE and reGATE; (2) inferring how the network connectivity structure varies with $y_i$; (3) assessing the predictive performance of the reGATE model. 

We train GATE and reGATE to obtain low-dimensional representations $z_i$'s.  Specifically, for the $n$ simulated networks, we have $\bar{z}_1=(\E(p_\theta(z_1\mid A_1)),\dots, \bar{z}_n = \E(p_\theta(z_n\mid A_n)))$ as the posterior means after training the GATE model. We then conduct principal component analysis (PCA) analysis on the posterior means $\{ \bar{z}_1,...,\bar{z}_n\}$ and plot each $\bar{z}_i$ using their first two PC scores in $\mathbb{R}^2$ colored according to the corresponding $y$ value. 
We can clearly observe the separation between these four types of networks in the low-dimensional representation space inferred using both GATE and reGATE, with reGATE yielding greater separation across the groups.

\begin{figure}[h!]
\centering
\includegraphics[width=6in]{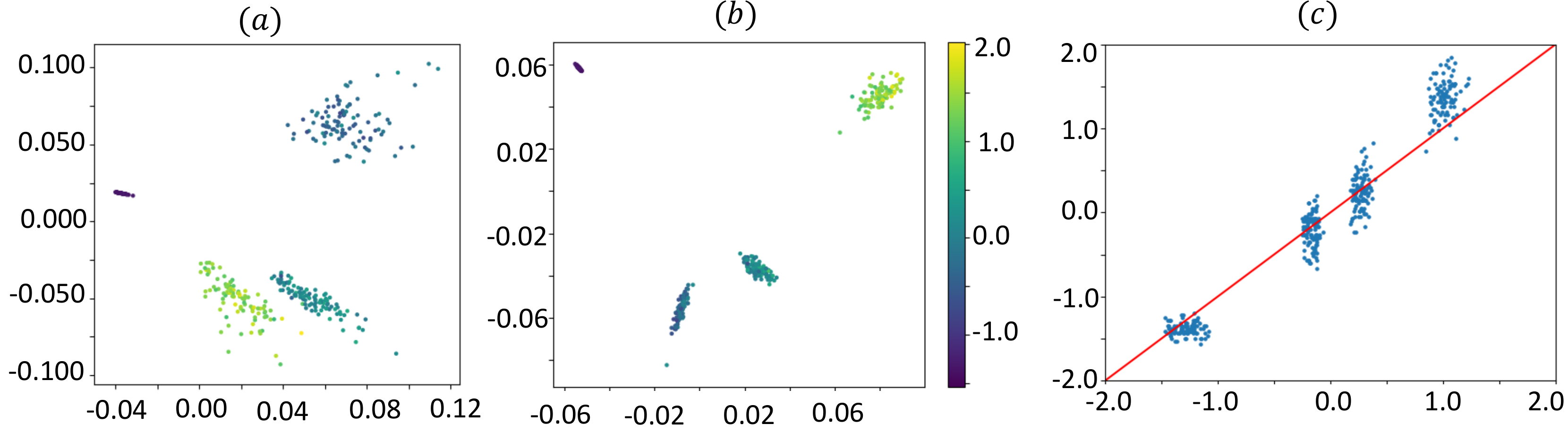}
\caption{\small Plot of the first 2 PC scores of the posterior mean of $z_i|A_i$ with corresponding $y_i$. Different colors refer to various values of  $y$. $(a)$ GATE, $(b)$ reGATE,  
$(c)$ Predicted $y$ v.s. true $y$ in reGATE, where x-axis is the true value  and y-axis is the predicted value.} \label{fig:sim:pca}
\end{figure}

To infer how brain networks $A_i$ vary according to $y_i$, we simulate networks from the posterior distribution of $A_i|y_i$ according to the conditional generative model in Section \ref{sec:regate}. Specifically, we first sample $z_i \mid y_i$ based on equation (\ref{eq:z_to_y:0}) with the parameters obtained from the previously trained reGATE and $y_i$ ranging from $-1.5$ to $2$, then generate networks via $p_\theta(A_i\mid z_i)$. 
Figure \ref{fig:sim:generate} shows the generated networks with structure varying for different $y_i$s. We can clearly observe that the network shows a sparse structure when $y_i=-1.5$, 
a mixture of community and small-world structure when $y_i=-0.1$, 
and a clear scale-free structure when $y_i=2$. These generated structures are consistent with the ground truth in our simulation settings. 

To address our third goal, we evaluate the predictive accuracy of reGATE. We consider two cases: (1) $y_i = \alpha^\top A_i \alpha + \epsilon_i$; (2) $y_i= (\alpha^\top A_i \alpha)^2 + (\alpha^\top A_i \alpha)^3 + \epsilon_i$. We also compare reGATE with a few popular methods in the literature for predicting human traits using network data. The first method is a regular linear regression based on tensor network principal component analysis (LR-TNPCA) in \cite{zhang2019tensor}.  The second method is linear regression based on a regular PCA applied to the vectorized networks. The third method is tensor regression proposed in \cite{zhou2013tensor}, denoted as CPR here. 
The mean square error (MSE) from five-fold cross-validation was used to compare different approaches. Figure \ref{fig:sim:pca} (c) shows the association between the predicted value and the true value under reGATE in case 1. The first and second rows in Table \ref{table:sim:mse} show the MSE under different methods. We can see reGATE outperforms other methods in predictive accuracy. 
The third row in Table \ref{table:sim:mse} reports the computing time with $100$ replicated simulations. reGATE is
both fast and accurate based on these results.

 \begin{table}[h!]
\centering
\begin{tabular}{p{2.5cm}p{2cm}p{2.5cm}p{2cm}p{2cm}}
\hline
 & reGATE  & LR-TNPCA & LR-PCA & CPR   \\
\hline
MSE: case 1 & {\bf 0.0252} &  0.0271 &  0.0340  & 0.0388   \\
MSE: case 2 & {\bf 0.0505} &  0.0593 &  0.0700  & 0.1078   \\
\hline
Time (mins) &  17.5   &   197.6  &  5.3   &  253.7    \\
\hline
\end{tabular}
\caption{The first row and the second row are the MSE under different methods. All numbers are calculated based on the mean of $100$ replicated simulations.}\label{table:sim:mse}
\end{table}

\begin{figure}[h]
\centering
\includegraphics[width=4in]{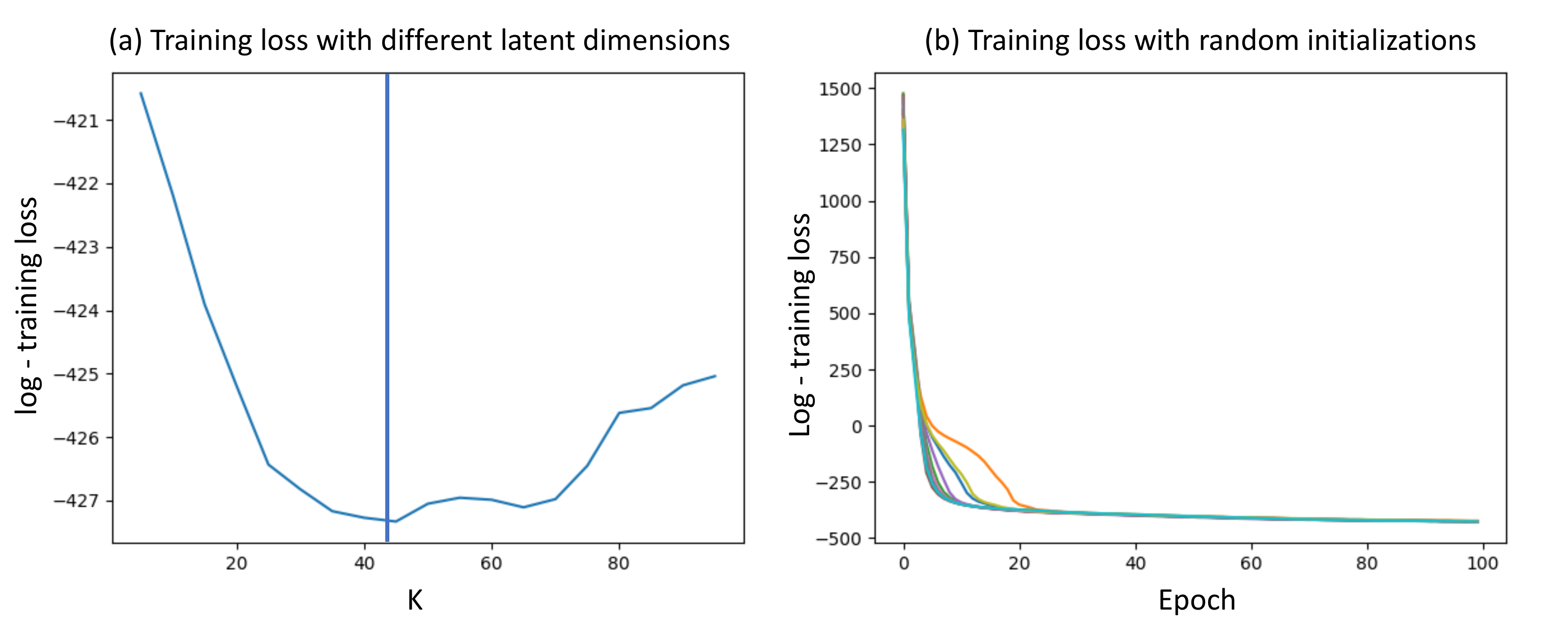}
\caption{\small (a) Log-training loss with latent dimension $K$ varies; (b) the paths of the log-training loss with random initializations. } \label{fig:sim:dim}
\end{figure}
We summarize the computing details used in the simulation study here. 
We run stochastic gradient descent with momentum (the Adam algorithm in \cite{kingma2014adam}) on GATE and reGATE with the learning rate $0.001$ on one NVIDIA Titan-V GPU. In GATE, we use a batch size of $128$, sampled uniformly at random at each epoch and repeated for $1000$ epochs. In reGATE, we use $5$-fold CV to calculate the MSE and run $200$ epochs with batch size of $128$ for each training dataset. The experimental details and network architectures for the inference model and generative model training are summarized in Table \ref{table:sim:design}. The latent dimension $K$ is chosen as the smallest value that achieves the minimal training loss, provided that the network architectures are fixed as in Table \ref{table:sim:design}. Figure \ref{fig:sim:dim} (a) shows the log-training loss with latent dimensions $K$ varying from $5$ to $100$, and the training loss achieves the best performance when $K$ is around $45$. The sensitivity of the training loss with respect to the random initialization is also explored. As shown in Figure \ref{fig:sim:dim} (b), under different initializations, the paths of the training loss converge to the same level as the number of epochs increases. 
All code used to produce the results and figures is available online via GitHub (https://github.com/meimeiliu/GATE).
\begin{table}[H]
\centering
\begin{tabular}{c|c|c|c|c}
\toprule
  & \multicolumn{2}{c|}{Inference model ($\mu_\phi$, $\sigma_\phi$)} & \multicolumn{2}{c}{Generative model} \\
\cline{2-5}
  & $\mu_\phi$ ($N=2$) & $\sigma_\phi$ ($N=2$)& setting & activation\\
\hline
\begin{tabular}[c]{@{}c@{}} GATE/reGATE \\ $(K=45)$ \end{tabular}& \begin{tabular}[c]{@{}c@{}}$W_{1,\mu}: 45*400 $\\$W_{2,\mu}: 400*45$\\$b_{1,\mu}:400*1$\\$b_{2,\mu}: 45*1$\\$\varphi_{1,\mu}= \mbox{ReLu}$\\$\varphi_{2,\mu}= \mbox{Linear}$ \end{tabular} &  
\begin{tabular}[c]{@{}c@{}} $W_{1,\sigma}: 45*400 $\\$W_{2,\sigma}: 400*45$\\$b_{1,\sigma}:400*1$\\$b_{2,\sigma}: 45*1$\\$\varphi_{1,\sigma}= \mbox{Relu}$\\$\varphi_{2,\sigma}= \mbox{Linear}$ \end{tabular} & \begin{tabular}[c]{@{}c@{}}  $k$-NN: 16 \\$M=2$ \\$R=5$ \end{tabular}
 &\begin{tabular}[c]{@{}c@{}}  $h_1$: Sigmoid \\$h_2$: Sigmoid  \end{tabular} \\
\bottomrule
\end{tabular}
\caption{\small Experimental details and network architectures. $K$ is the dimension of $z_i$, $N$ is the number of layers in the inference network, $M$ is the number of layers in GCN, and $R$ is the dimension of $X^{(i)}$. } \label{table:sim:design}
\end{table}

\section{Applications to the ABCD and HCP data}\label{sec:real}

We apply our method to both the ABCD and HCP datasets described in Section \ref{subsec:data} to examine the relationship between structural brain networks and cognition for adolescents and young adults. The  
ABCD study uses a reliable and well-validated battery of measures that assess a wide range of human functions, including cognition. The core of this battery is comprised of the tools and methods developed by the NIH Toolbox for assessment of neurological and behavioral function \cite{GershonS2}. The Toolbox includes measures of cognitive, emotional, motor, and sensory processes. Since we are particularly interested in cognition, we extract four cognition related measures as $y$ from ABCD, including
\begin{enumerate}[$(a)$]
    \item Picture vocabulary test: the picture vocabulary test uses an audio recording of words, presented with four photographic images on the computer screen. Participants are asked to select the picture that best matches the meaning of the word. 
    \item Oral reading recognition test: participants on this test are asked to read and pronounce letters and words as accurately as possible. 
    \item Crystallized composite score: crystallized cognition composite can be interpreted as a global assessment of verbal reasoning.  We use the age-corrected standard score. 
    \item The cognition total composite score: this composite score measures overall cognition and is obtained from a factor analysis \cite{heaton2014reliability}.
\end{enumerate}
We extracted two matched cognitive measures from HCP: picture vocabulary test score and oral reading recognition test. In addition, we add two more cognitive measures in our data analysis: total number correct answers and total positions off in a line orientation test.  A more detailed description of these traits can be found in \cite{GershonS2}.  

\subsection{Visualization: show network data  in low-dimensional space}\label{subsec:abcd:visua}
\begin{figure}[htb!]
 \centering
 \begin{tabular}{cc}  
\includegraphics[scale=0.28]{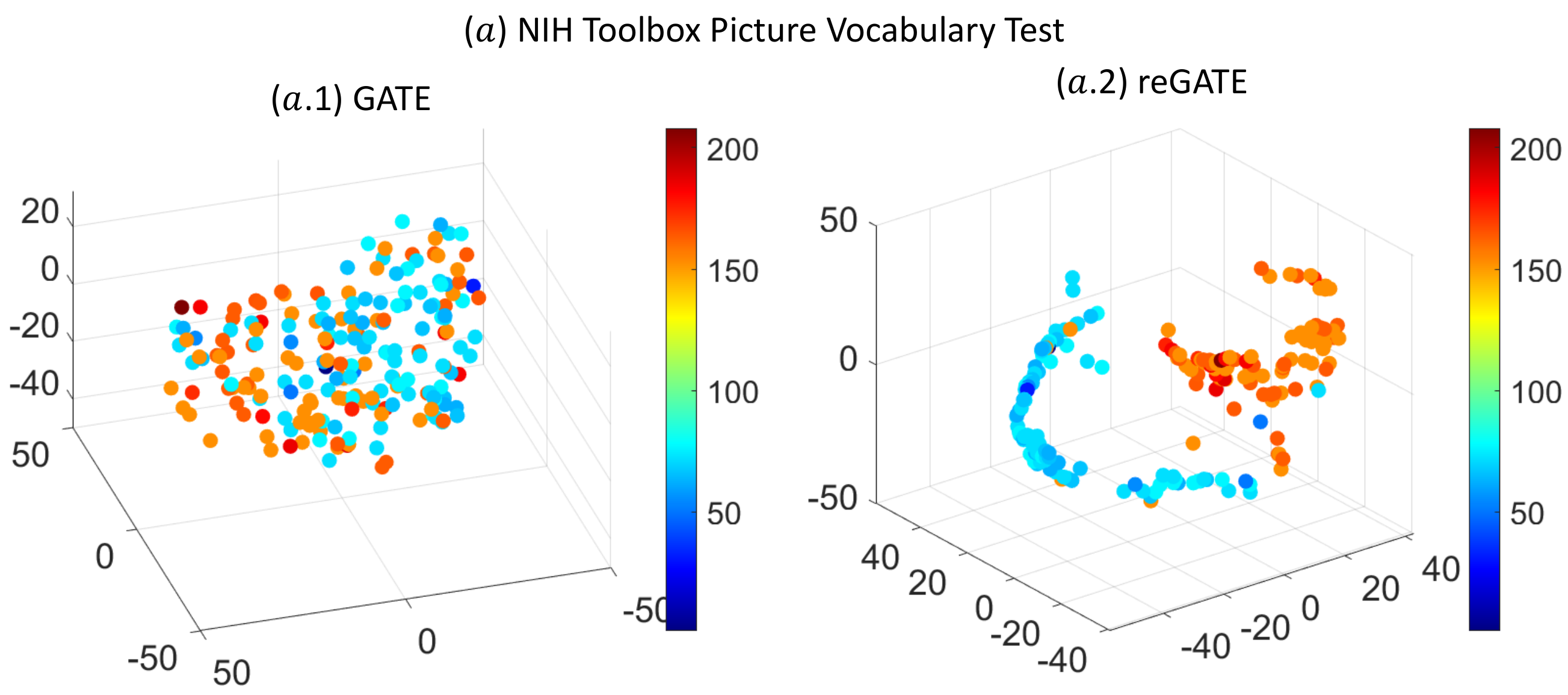}&
\includegraphics[scale=0.28]{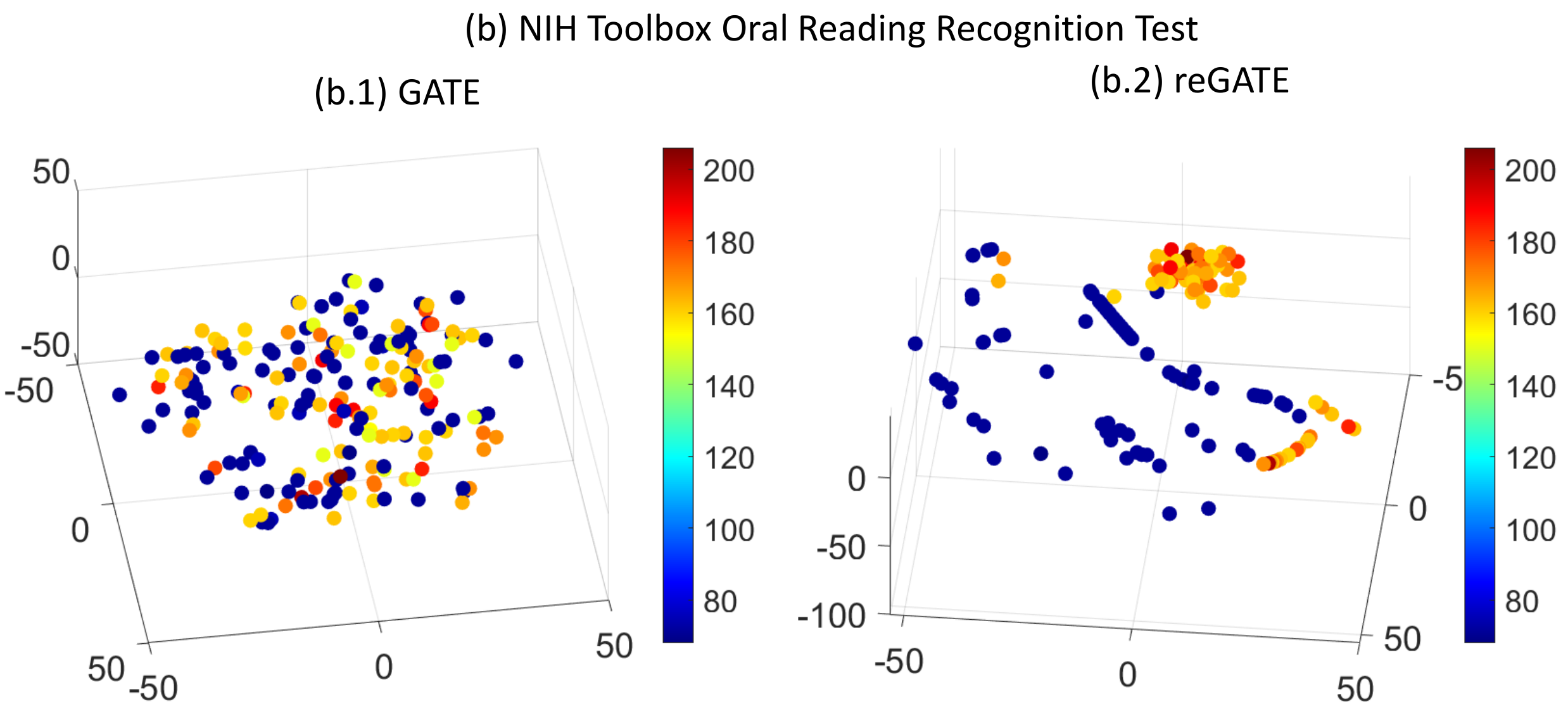} \\  
\includegraphics[scale=0.28]{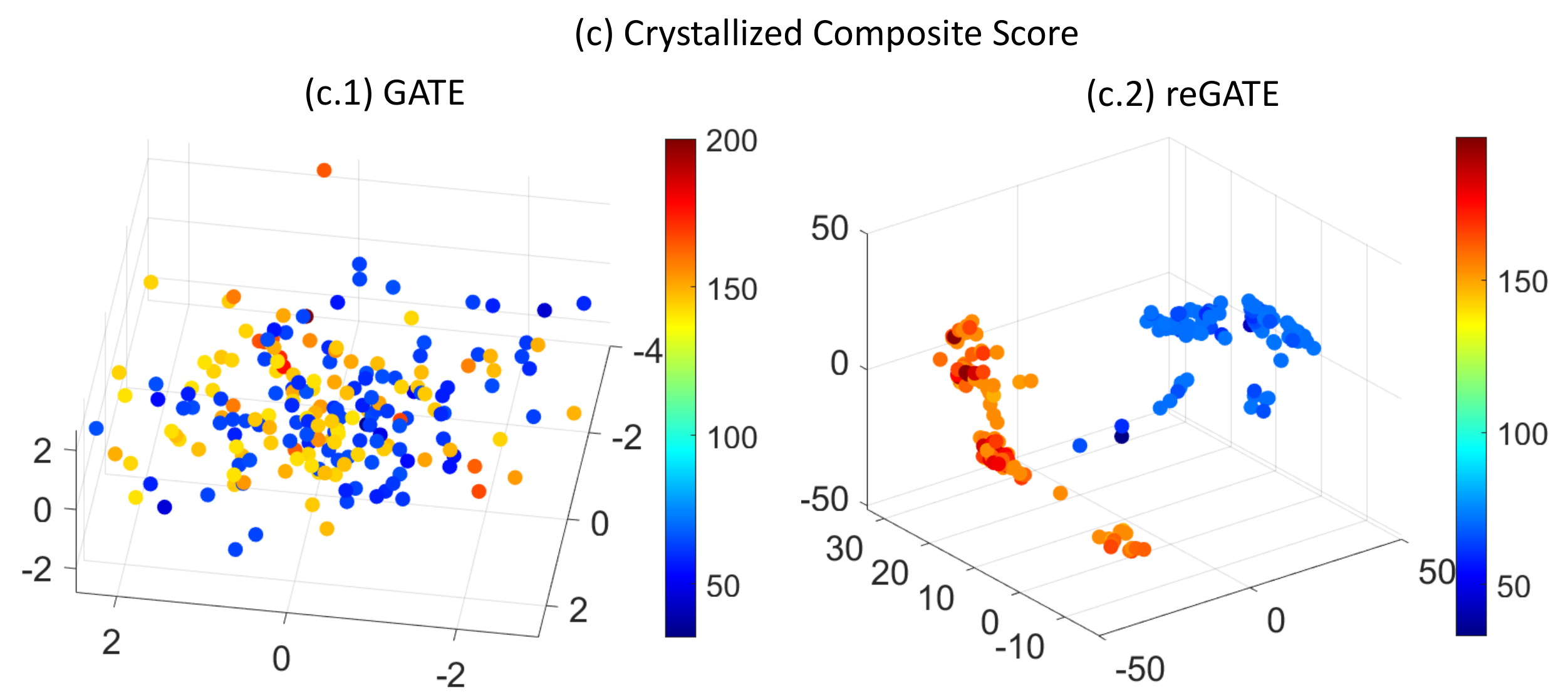}&
\includegraphics[scale=0.28]{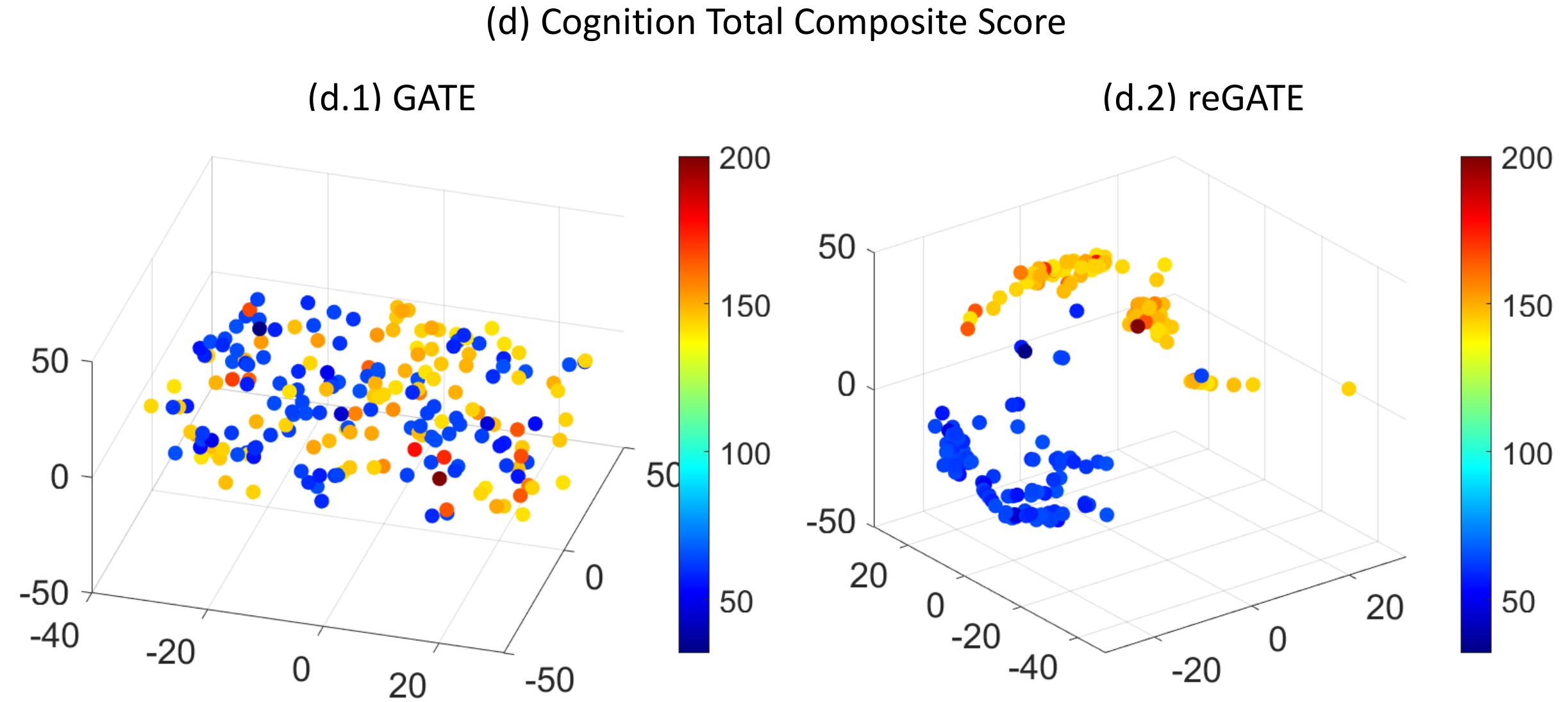} \\   

 \end{tabular}
  \caption{\small Visualization of low dimensional embedding $z_i$ learned from GATE and reGATE applied to the ABCD study. We display $100$ subjects with the lowest trait scores and $100$ subjects with the highest scores for each trait. Colors represent the trait scores. 
   } \label{fig:real:abcd:clustering}
\end{figure}

Both GATE and reGATE output low-dimensional representations of the brain networks. We can visualize the latent features of each individual's connectome and examine the relationship between structural connectivity and the four traits via the latent features. We train GATE on $5252$ brain networks extracted from the ABCD dataset to obtain low-dimensional representations $z_i$ for $i=1,\dots,5252$. We then plot the posterior mean of $z_i\mid A_i$ using t-SNE \cite{van2008visualizing} colored with its corresponding trait score in $\mathbb{R}^3$. 
We show $200$ subjects' data for each cognition trait, with the first $100$ subjects having the lowest trait scores and the second $100$ subjects having the highest scores. As shown in Figure  \ref{fig:real:abcd:clustering} $(a)-(d)$, under both GATE and reGATE, we obtain a large separation between the two groups of subjects, indicating that brain connection patterns are different for these groups. reGATE has better performance since we incorporate the trait information in learning the $z_i$s.  A similar analysis is conducted using the HCP data, and the result is shown in Figure \ref{fig:hcp:clustering}  in the supplement.

\subsection{Prediction: predict traits with reGATE}
In this section, we demonstrate the predictive ability of reGATE and compare it with several competitors: LR-TNPCA \cite{zhang2019tensor}, LR-PCA, CPR \cite{zhou2013tensor}, and BLR \cite{wang2019symmetric}. BLR is supervised bi-linear regression (BLR) with emphasis on signal sub-network selection. To verify the role played by the latent space model, 
we further compare our approach with a simplified reGATE without considering the node-level latent space structure. Specifically,  we simplify the generative model as 
$
z_i \sim  \; N(0, I_K)$, and 
$A_{i\ell}| z_i \sim  \; \textrm{Poisson}\big(\lambda_{il}(z_i)\big) , 
$ where $\lambda_{il}(z_i)$ is directly learned via a fully connected neural network without the latent space setting in equations (\ref{eq:decompose_1})-(\ref{eq:decompose_3}). We denote this approach as S-reGATE. 

MSE from five-fold cross-validation is used to assess performance. 
Table \ref{table:abcd:mse:real} shows results for ABCD and HCP datasets. reGATE significantly outperforms other methods in both datasets for all traits. S-reGATE has a better prediction performance in ABCD data but cannot compete with BLR in the HCP dataset. Hence, the proposed latent space model does indeed improve prediction; interpretation is also improved relative to S-reGATE, which does not capture some aspects of variation in the brain networks. 

With the bigger sample size ($n=5252$) in the ABCD data, the improvements of reGATE are more significant than those for the HCP data. It is well known that deep neural networks tend to perform exceptionally well for large training sample sizes.

\begin{figure}[htb!]
 \centering
\includegraphics[height=1.6in]{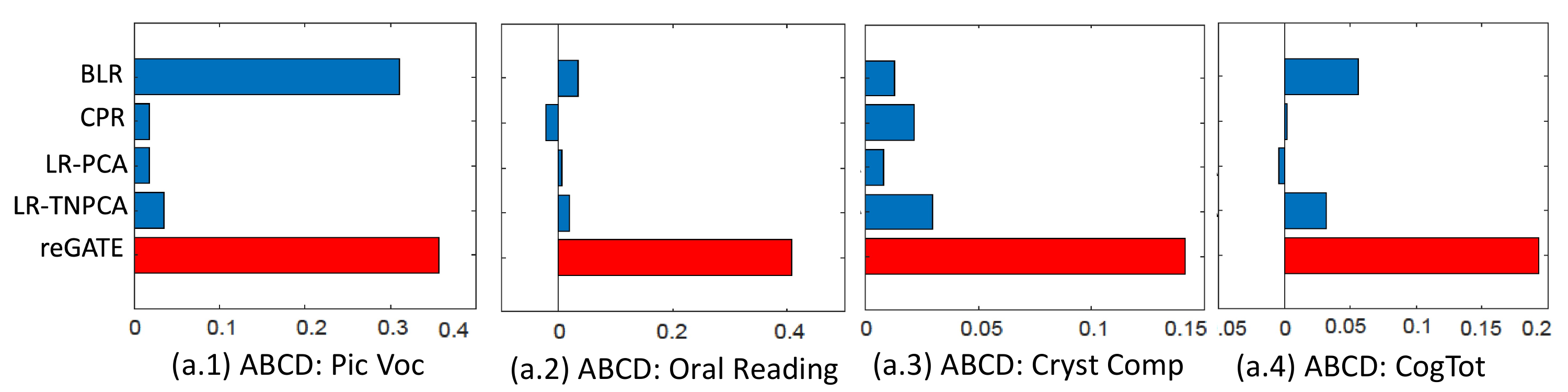}\\
\includegraphics[height=1.48in]{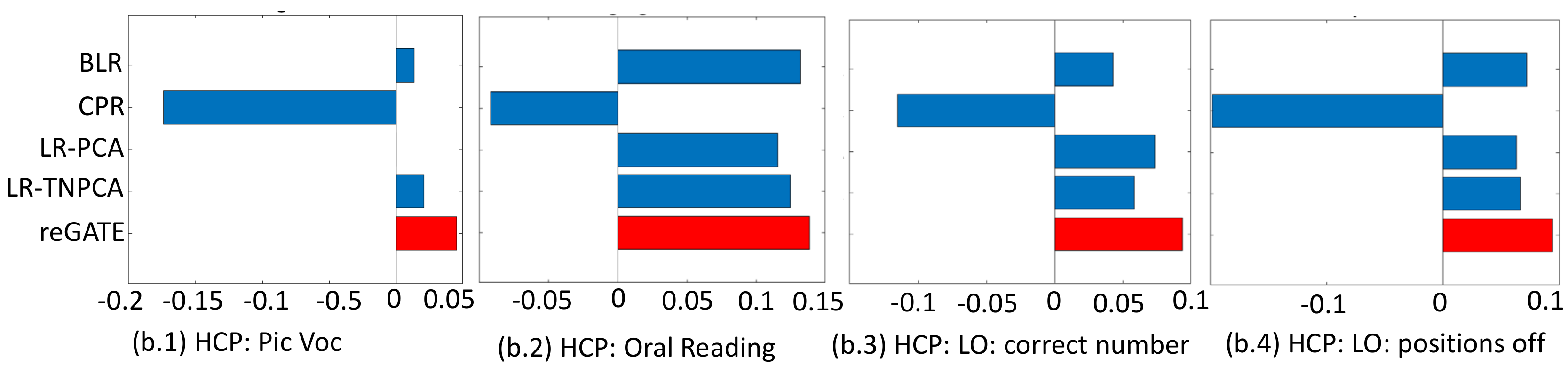}\\
  \caption{\small Percentage of the MSE improvement compared with a baseline method of using $\bar{y}$ as the predicted trait. (a.1)-(a.4): ABCD with Picture vocabulary score, oral reading recognition test score, crystallized composite age-corrected standard score, and cognition total composite score. (b.1)-(b.4): HCP with Picture vocabulary score, oral reading recognition test score, LO-correct number, LO-positions off. 
  The X-axis is the improved proportion; the y-axis marks different methods. } 
  \label{fig:real:abcd:traits_improve}
\end{figure}

We also calculate the percentage of MSE improvement upon a baseline model, the sample means $\bar{y}$, for the prediction result from the ABCD and HCP data. As shown in Figure \ref{fig:real:abcd:traits_improve} (a.1)-(a.4), most methods in ABCD study demonstrate an improved MSE compared with the sample mean $\bar{y}$, indicating that there is a detectable relationship between the structural connectome and cognitive traits. However, reGATE achieves stable and significant prediction improvements ranging from $30\%$ to $40\%$, while the competitors' performance improvements fluctuate from $-5\%$ to $5\%$. From Figure \ref{fig:real:abcd:traits_improve} (b.1)-(b.4), we see reGATE still has the best performance among other competitors.

Besides MSE, we evaluate the correlation between the predicted value $\widehat{y}_i$ and the observed value $y_i$ via five-fold cross-validation for different approaches. Correlation is reported in the parenthesis in Table \ref{table:abcd:mse:real}. We can see that reGATE significantly improves the prediction by increasing the correlation from around $0.2$ to $0.4$.

\begin{table}[h!]
\small
\centering
\begin{tabular}{@{}rllllll@{}}
\toprule
 & reGATE& S-reGATE & LR-TNPCA & LR-PCA & CPR & BLR  \\
\hline
ABCD ($n=5252$) \\
{\small Pic Voc} &  {\bf 186.8 (0.40)}  &212.5 (0.27) &280.1 (0.22) &  285.2 (0.20) & 285.1 (0.23) & 200.4 (0.29)
\\
{\small  Oral Reading} & {\bf 206.5 (0.41)} & 224.8 (0.32)   & 342.3 (0.19) &  346.5 (0.14) & 357.0 (0.13) & 337.1 (0.19) \\ 
{\small Cryst Comp} & {\bf 279.4 (0.39)}  &  304.5 (0.31) & 315.8 (0.23) &  322.9 (0.22) & 318.4 (0.19) & 321.3 (0.30) \\  
{\small CogTot Comp} & {\bf 248.0 (0.38)} &  255.6 (0.35) &297.6 (0.17)  &  308.6 (0.26) & 306.7 (0.22) & 290.2 (0.32) \\ 
\hline
HCP ($n=1065$)\\
{\small Pic Voc} & {\bf 209.1 (0.28)} & 214.4 (0.25) & 214.5 (0.25) & 219.0 (0.21) & 257.2 (0.17) & 216.1 (0.23)\\ 
{\small Oral Reading} &  {\bf 198.9 (0.26)} & 205.7 (0.22)  & 202.1 (0.22) & 204.2 (0.21) & 252.1 (0.19) & 200.4 (0.26)
\\
{\small LO: correct number} &{\bf \, 17.8 (0.27)}  & \,20.4 (0.20) &\, 18.5 (0.23) &\, 18.2 (0.22) &\, 21.9 (0.19) &\, 18.8 (0.18) \\ 
{\small LO: positions off} & {\bf 195.8 (0.27)} & 203.5 (0.25) & 201.8 (0.22) & 202.6 (0.24) & 259.0 (0.19) & 200.7 (0.27)\\ 
\bottomrule
\end{tabular}
\caption{Comparison of prediction results of different methods in both the ABCD and HCP datasets. Each cell shows the MSE and correlation (in the parenthesis) between the observed and predicted measures via five-fold CV.} \label{table:abcd:mse:real}
\end{table}

\subsection{Inference: generate brain connectomes conditioned on traits}\label{subsec:abcd:generate}
\begin{figure}[htbp]
 \centering
 \includegraphics[width=6in]{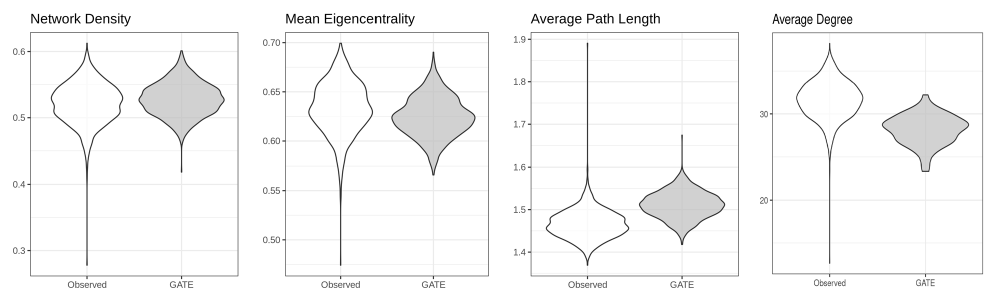}
\caption{ \small HCP: Goodness-of-fit assessments for selected network summary measures. The violin plots summarize the distribution in the observed data (white) and the posterior predictive distribution arising from GATE (gray). 
 } \label{fig:real:density}
\end{figure}

An appealing characteristic of reGATE is the ability to 
generate brain networks for new individuals conditionally on their trait value.
This facilitates inference on 
how brain networks, and their topological properties, change across different levels of a trait and how this dependence varies across individuals. In our first experiment, we assess the performance of GATE in characterizing the observed brain network data via posterior predictive checks \cite{gelman2013bayesian} for relevant network topological properties, including network density, mean eigencentrality, average path length, and average degree. Denote $\eta_k = g_k\{L(\mathcal{A})\}$ as the random variable associated with the $k$th network summary measure, for  $k=1,2,3,4$, where $\mathcal{A}$ represents the brain connectomes. Then we calculate the posterior predictive distributions for  
these summary measures based on 
the generative model $p_\theta(\mathcal{A}\mid z)$ in Section \ref{sec:generative} as
$
p_{\eta_k}(\eta \mid z) = \sum_{a\in \mathbb{A}_v:g_k(a)=\eta} p_{\theta}(\mathcal{A}=a \mid z),\; \textrm{for}\; k=1,2,3, 4, 
$
respectively. Figure \ref{fig:real:density} compares the network summary measures computed using all 1065
subjects from the HCP data (white-colored) and the generated network data (gray-colored) from GATE. GATE achieves good performance in characterizing the observed network summary measures.

\begin{figure}[h!]
\centering
\includegraphics[scale=0.5]{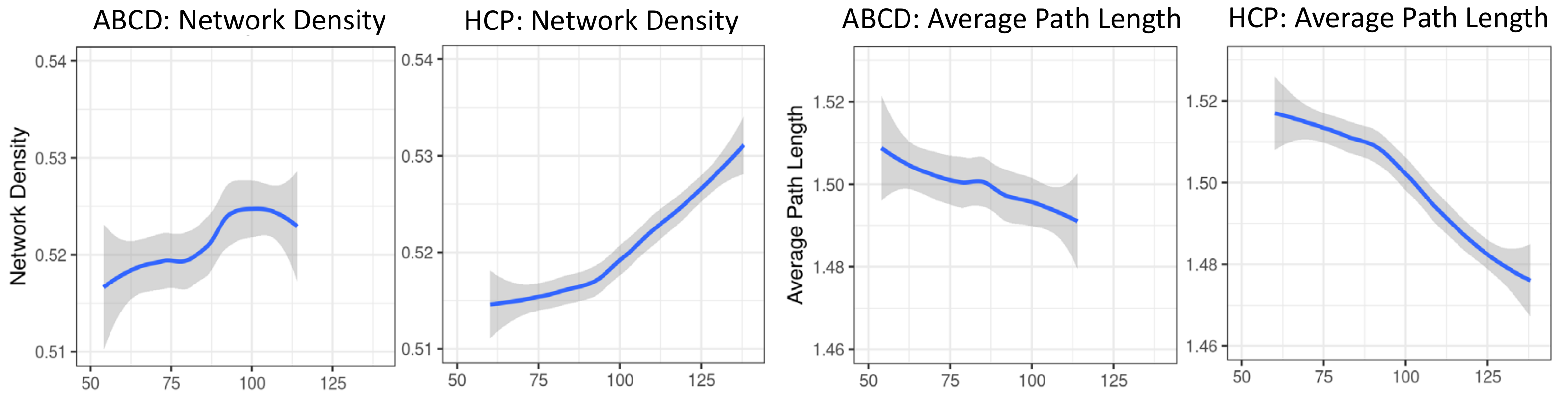}
\caption{\small ABCD and HCP: posterior predictive network summaries with confidence bands for network density and average path length. The X-axis is the picture vocabulary test score. The confidence bands are calculated based on $500$ generated network summaries conditional on different picture vocabulary test scores. 
 } \label{fig:confband}
\end{figure}

We next consider the conditional predictive distribution of network topological summaries given trait scores, i.e., $\eta_k \mid y$. We consider network density and average path length for both the ABCD and HCP data and focus on different levels of picture vocabulary test score as $y$. 
The distribution of $\eta_k \mid y$ can be expressed as 
$
p_{\eta_k}(\eta \mid y) = \sum_{a\in \mathbb{A}_v:g_k(a)=\eta} p_{\theta}(\mathcal{A}=a \mid y), \; \textrm{for}\; k=1, 2,  
$
respectively, where $p_{\theta}(\mathcal{A}=a \mid y)$ can be obtained via the conditional generative model in Section \ref{sec:regate}. More specifically, conditional on each particular level of picture vocabulary test score, we first generate 500 networks based on the well-trained reGATE model. Next, we calculate the network summaries for each generated network, and then calculate the mean and confidence bands based on 500 calculated network summaries. As shown in Figure \ref{fig:confband}, the network density increases and the average path length decreases as the picture vocabulary score increases. There is more variability in network topological summaries for the adolescent subjects in ABCD than for the adults in HCP.

\begin{figure}
 \centering
\begin{tabular}{c}
\includegraphics[height=3.5in]{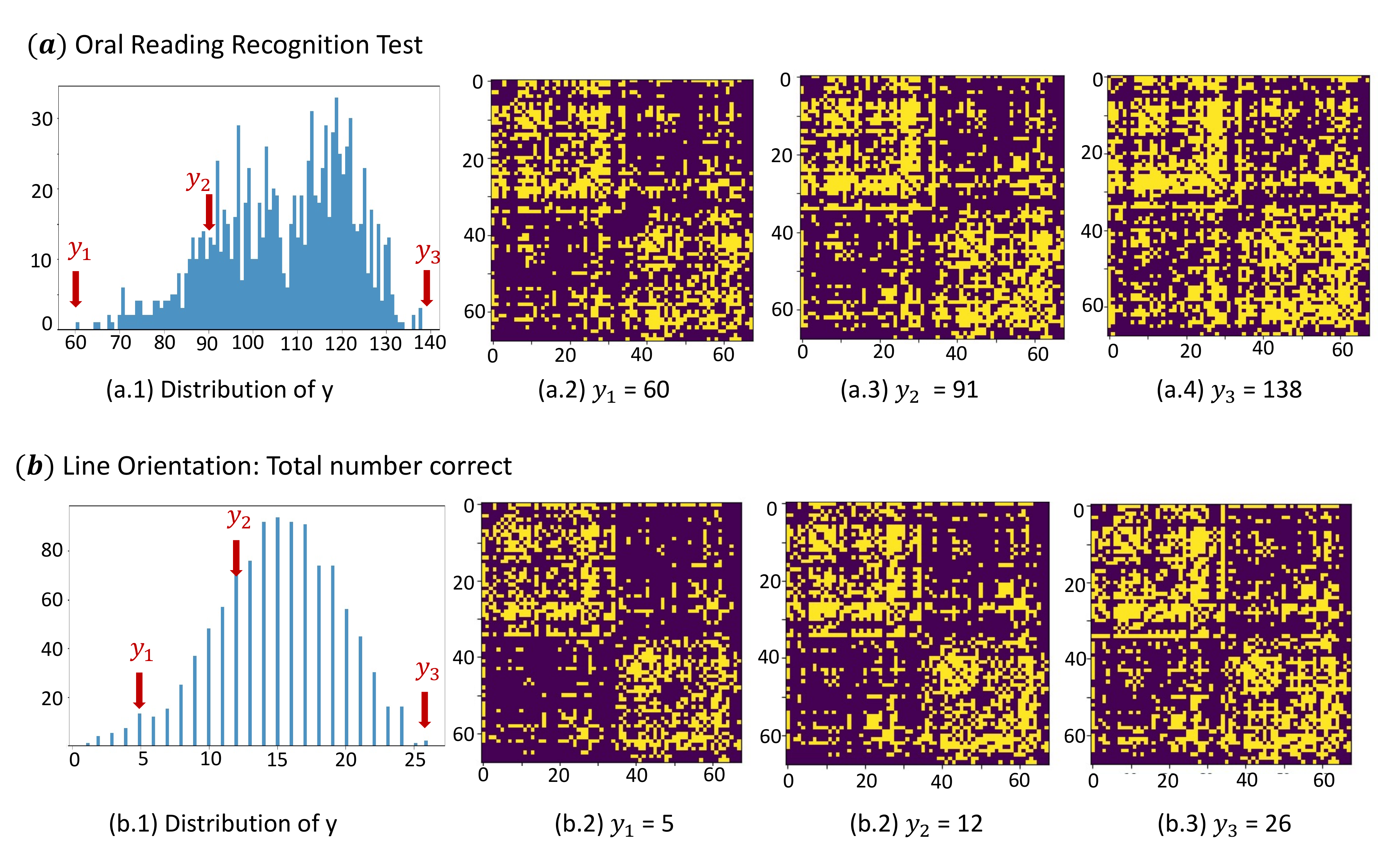}
 \end{tabular}
\caption{ \small Examples on generating conditional brain networks $A\mid y$ for different $y$. The first column shows histograms of observed trait scores for the $1065$ individuals in HCP (the $x$-axis shows the trait value, and the $y$-axis shows the frequency). The second to the fourth columns show the mean of generated brain networks (after dichotomization) conditional on different $y$ under the reGATE model (with the $x$-axis and the $y$-axis indexing brain regions).
   } \label{fig:generate:diff}
\end{figure}

We next explore how the brain network varies across different levels of trait $y$.  We consider three levels that range from the minimum to the maximum for each trait and generate multiple networks for each trait value. For example, for the oral reading recognition score in the HCP, we consider $y=60, 91$ and $138$, and generate brain networks according to the conditional generative procedure in Section \ref{sec:regate}. If we set $y=60$, then the posterior $p_\theta(A\mid y=60)$ indicates the distribution of the brain networks for people with oral reading recognition test equal to $60$.   
A mean network is used to summarize the generated network data for a given $y$, and we further dichotomize the connectome to $\{0, 1\}$ depending on whether a connection exists for better visualization. The results are shown in Figure \ref{fig:generate:diff}, where the first column shows histograms of the observed trait scores, and the second to the fourth columns show the generated mean networks (with the first 34 nodes from the left side of the brain and the next 34 nodes from the right side of the brain; The Supplement contains a table with descriptions of each ROI) for different $y$. From the result in the first row, more connections between the brain's two hemispheres are correlated with better reading ability. In the second row, we observe the same pattern for the line orientation trait. This result is consistent with findings in the literature \cite{bullmore2009complex, durante2018bayesian}. 
 
\begin{figure}[h!]
\centering
 \begin{tabular}{cc}
\hline
\multicolumn{2}{l}{HCP: picture vocabulary reading  scores} \\
\vspace{3mm}
\includegraphics[scale=0.25]{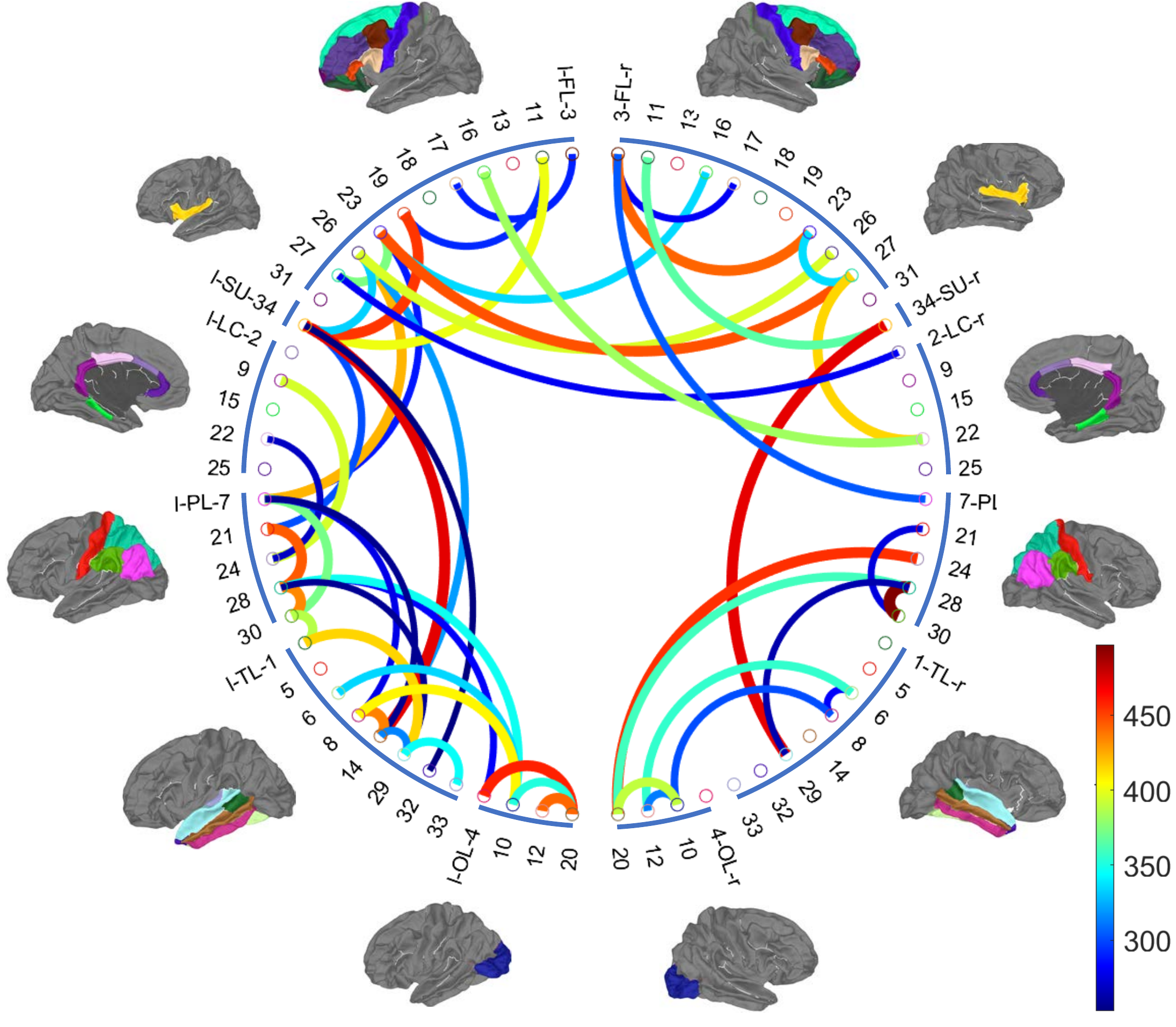}  &
\includegraphics[scale=0.25]{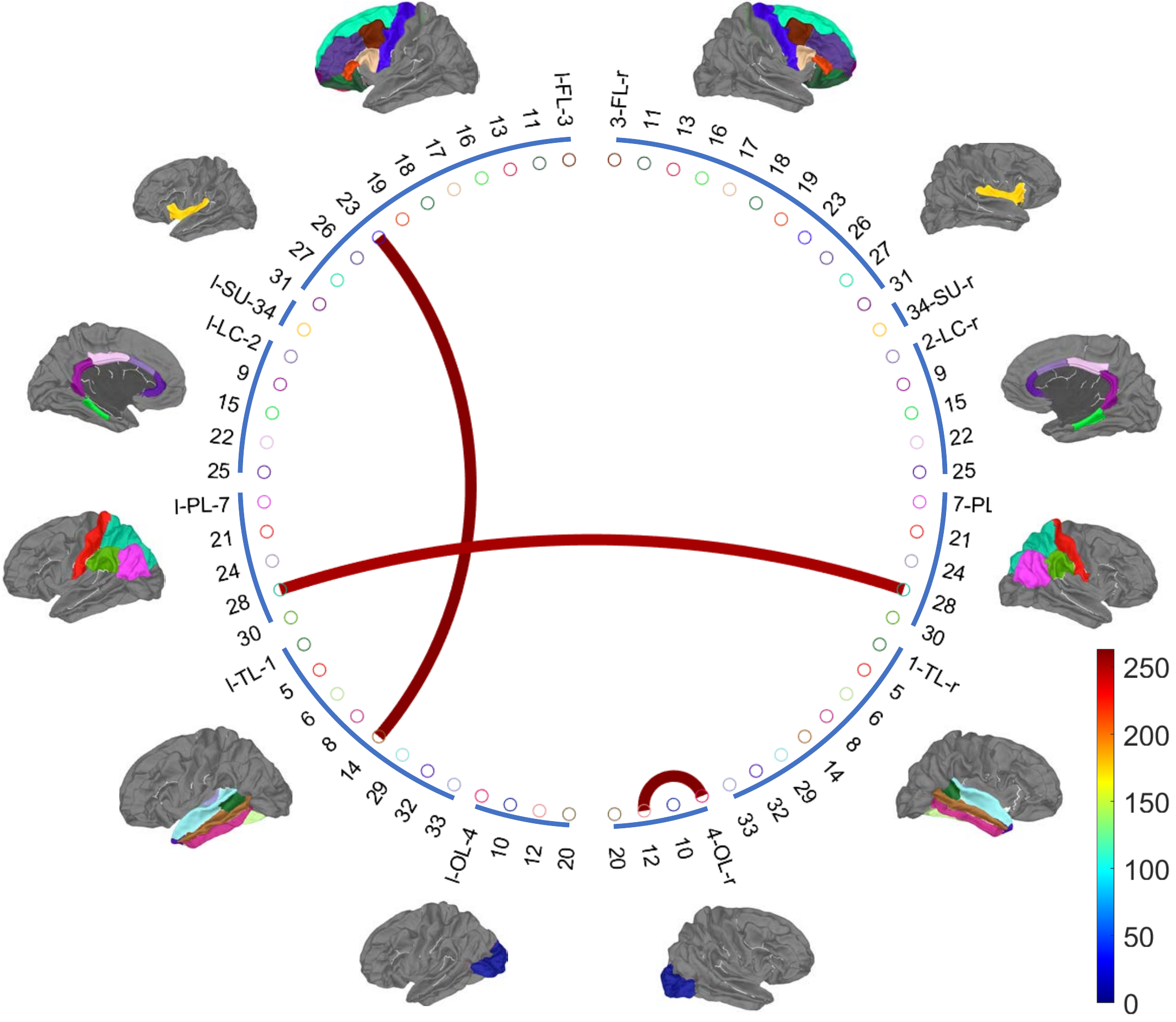}\\ 
\hline
\multicolumn{2}{l}{ABCD: picture vocabulary  reading scores} \\
\vspace{3mm}
 \includegraphics[scale=0.25]{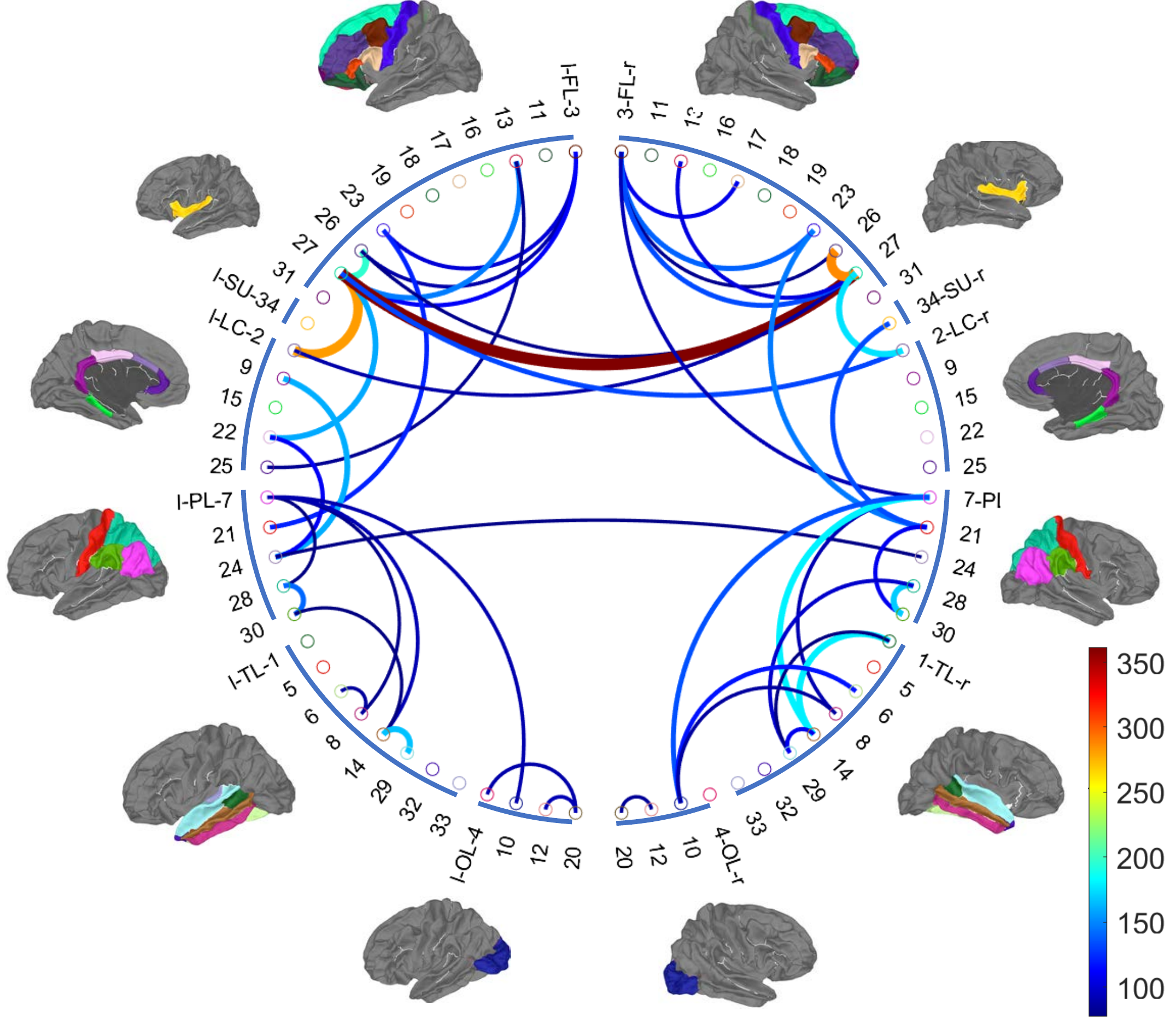}&
 \includegraphics[scale=0.25]{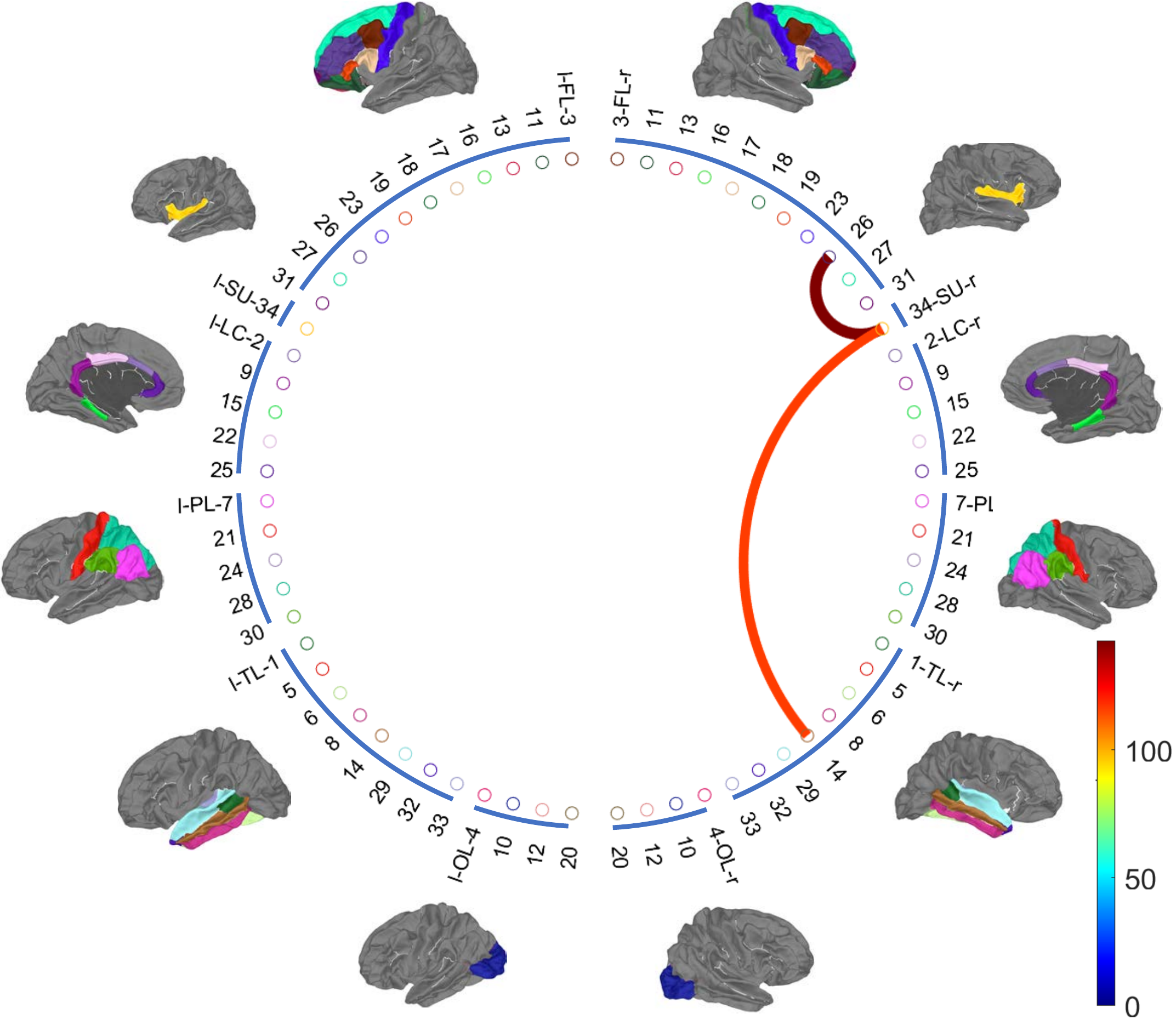} \\
 \hline
 \end{tabular}
\caption{\small Top $50$ pairs of brain regions in terms of the mean changes in the generated brain connectivity between two levels of picture  Top $50$ pairs of brain regions in terms of the mean changes in the generated brain connectivity between two levels of picture vocabulary reading scores ($90$\% and $10$\% quantiles).
The left panel shows the positive connections among the 50 pairs, and the right panel shows negative connections. More details about each node can be found in the Excel spreadsheet in Supplementary Material II.  vocabulary reading scores ($90$\% and $10$\% quantiles).
The left panel shows the positive connections among the 50 pairs, and the right panel shows negative connections. More details about each node can be found in the Excel spreadsheet in Supplementary Material II. 
 } \label{fig:meandiff}
\end{figure}

To enhance the above analysis, we reduce the number of levels and select only the $10\%$th and 
$90\%$th quantiles from the observed $y$ as two representative levels. For each level, we generate $100$ conditional networks $A_i \mid y_i$ and calculate the mean difference between the networks in the high and low trait groups. Figure \ref{fig:meandiff} plots the top $50$ connections (based on absolute values) in the mean difference network for the picture vocabulary test for both HCP and ABCD. This procedure is done separately for the two datasets. The $50$ connections are further separated into positive and negative connections and are plotted in separate panels in Figure \ref{fig:meandiff}. After ranking the connections by absolute values,
these connections are dominated by positive ones, indicating that better vocabulary ability is associated with more connections.

The first row in Figure \ref{fig:meandiff} shows results from the HCP data, and the second row shows results from the ABCD data. Considering the different populations (young adults vs adolescents) in our data analysis, it is interesting to observe many
similar results.
For example, we observe denser connections both within the left and right frontal lobes and between them.  In particular, we see that brain regions such as  $l26,r26$ (rostral middle frontal), $l27,r27$ (superior frontal), and $l3,r3$ (caudal middle frontal) are densely involved (nodes with high degrees) in the top 50 connections. These brain regions are thought to contribute to higher cognition and particularly working memory \cite{boisgueheneuc2006functions}, and are important for language-related activities \cite{binder1997human,friederici2002towards}. We also observe that the four nodes in the occipital lobe (regions $4,10,12$ and $20$) all appear in the top 50 connections. This visual preprocessing center has a few connections (fiber bundles) to the parietal lobe (e.g., regions $7,24,28$) and then to the frontal lobe. Hence, our results show that richer connections between the visual, sensory, and working memory systems are strongly associated with higher picture vocabulary in adolescents and adults. Some negative connections within each hemisphere are also observed, although their numbers and strengths are smaller than the positive ones.  These connections may arise from errors introduced during the connection recovery stage or simply statistical noise.

\subsection{Computing details}
We summarize the computing details for our analyses of ABCD and HCP data. We run Adam on GATE and reGATE with the learning rate $0.001$ on one NVIDIA Titan-V GPU. In GATE, we used a batch size of $128$ sampled uniformly at random at each epoch and ran $200$ epochs. In reGATE, we used $5$-fold CV to calculate the MSE and ran $100$ epochs with a batch size of $128$ for each training dataset. The latent dimension $K$ can be chosen as $68$ under the same criteria as in Section \ref{sec:sim}. The computing time in modeling the ABCD dataset is 14.34 mins for GATE and 15.2 mins for reGATE. The computing time in the HCP dataset is 4.74 mins for GATE and 5.18 mins for reGATE. 

Table \ref{table:real:design} shows the detailed network architectures for the inference model and generative model training.  

\begin{table}[H]
\centering
\begin{tabular}{c|c|c|c|c}
\hline
  & \multicolumn{2}{c|}{Inference model ($\mu_\phi$, $\sigma_\phi$)} & \multicolumn{2}{c}{Generative model} \\
\cline{2-5}
  & $\mu_\phi$ ($N=2$) & $\sigma_\phi$ ($N=2$)& setting & activation\\
\hline
\begin{tabular}[c]{@{}c@{}} GATE/reGATE \\ $(K=68)$ \end{tabular}& \begin{tabular}[c]{@{}c@{}}$W_{1,\mu}: 68*256 $\\$W_{2,\mu}: 256*68$\\$b_{1,\mu}:256*1$\\$b_{2,\mu}: 68*1$\\$\varphi_{1,\mu}=$Relu\\$\varphi_{2,\mu}=$Linear \end{tabular} &  
\begin{tabular}[c]{@{}c@{}} $W_{1,\sigma}: 68*256 $\\$W_{2,\sigma}: 256*68$\\$b_{1,\sigma}:256*1$\\$b_{2,\sigma}: 68*1$\\$\varphi_{1,\sigma}=$ Relu\\$\varphi_{2,\sigma}=$Linear \end{tabular} & \begin{tabular}[c]{@{}c@{}}  $k$-NN: 32\\$M=2$ \\$R=5$ \end{tabular}
 &\begin{tabular}[c]{@{}c@{}}  $h_1$: Sigmoid \\$h_2$: Sigmoid \end{tabular} \\
\hline
\end{tabular}
\caption{\small Experimental details and network architectures. $K$ is the dimension of $z_i$, $N$ is the number of layers in the inference network, $M$ is the number of layers in GCN, $R$ is the dimension of $X^{(i)}$. } \label{table:real:design}
\end{table}

\section{Discussion}
We develop a novel nonlinear latent factor model to characterize the population distribution of brain connectomes across individuals and depending on human traits. GATE outputs two layers of low dimensional nonlinear latent representations: one on the individuals that can be used as a summary score for visualization and prediction of human traits of interest; and one on the nodes for characterizing the network structure of each individual based on a latent space model. A supervised model reGATE is proposed to analyze the relationship between human traits and brain connectomes. 
GATE/reGATE are developed based on a deep neural network framework and implemented via a stochastic variational Bayesian algorithm. The algorithm is computationally efficient and can be applied to massive networks with large number of nodes (brain ROIs). 

With applications to the ABCD and HCP data, we used GATE and reGATE to study the relationship between brain structural connectomes and various cognition measures. Using the generative model of reGATE, we can simulate brain networks for a given $y$ (e.g., cognition measure) and compare these brain connectomes and related network topological summaries under different cognition levels while allowing variability across individuals. For these cognition traits, we clearly observe that cross hemispheres connections are essential. In the ABCD and HCP datasets, we found that the network density increases while the average path length decreases as the cognition level increases; these network measures for adolescents evaluated in ABCD have a higher variation than adults in HCP. 
reGATE had superior performance in predicting trait scores from brain networks, with the gain particularly notable in the larger ABCD study. 
  
The generative aspect of GATE/reGATE has a wide range of applications, including data augmentation, outlier network detection, and potentially sensitive data release. As an example in data augmentation, most existing neuroimaging studies contain only a few subjects, and models like GATE that require a large sample size do not work well for such data. With the help of large datasets, such as the ABCD and UKBiobank data, we can pre-train a model using the large datasets and then refine the model with the smaller dataset. With our generative model, we can generate more data to mimic the data distribution of the smaller dataset and release confidential datasets generated from this distribution.
  
In the future, we would like to extent GATE/reGATE in the following directions. First, in brain networks, a refined brain division can give a larger number of nodes, providing a more detailed description of the brain. GATE and reGATE provide a new set of tools for handling high-resolution brain networks, and it becomes interesting to extend the methodology to handle multiresolution data. Moreover, current large studies all collect both functional MRI and diffusion MRI data. It is straightforward to extend GATE/reGATE to jointly embed both functional and structural connectomes, even allowing the strength and nature of the link to vary across individuals and with traits.

\bibliographystyle{plain}
\bibliography{ref}

\newpage
\clearpage
\setcounter{page}{1}
\setcounter{section}{0}
\renewcommand{\thesection}{S.\arabic{section}}
\setcounter{subsection}{0}
\renewcommand{\thesubsection}{S.\arabic{subsection}}
\setcounter{equation}{0}
\renewcommand{\theequation}{S.\arabic{equation}}
\onecolumn

\begin{center}
  \begin{center}
  {\large\bf SUPPLEMENTARY MATERIAL\\
Auto-encoding brain networks with application to analyzing large-scale brain imaging datasets}
  \end{center}
\end{center}

This Supplement lists the detailed neural network modeling framework for GATE/reGATE and describes additional numerical studies.  

\subsection{Parameter setup in $q_{\phi}(z_i\mid A_i)$ defined in Equation (\ref{eq:q}) for GATE }\label{subsec:appendix:q}
In Section \ref{subsec:gate_learning}, we aim to approximate $p_\theta(z_i\mid A_i)$ by  $q_{\phi}(z_i\mid A_i)$, where $q_{\phi}(z_i\mid A_i)\sim N(\mu_\phi, \sigma_\phi^2)$. 
Particularly, we learn $\mu_\phi$ and $\sigma_\phi^2$ via 
\begin{eqnarray}
\mu_\phi(A_i) = & \varphi_{N,\mu}[W_{N,\mu}\varphi_{N-1,\mu}\{W_{N-1,\mu}\cdots\varphi_{1,\mu}(W_{1,\mu} A_i+b_{1\mu})+b_{N-1,\mu}\} + b_{N,\mu}], \label{eq:inf:mu}\\
\diag\{\sigma^2_\phi (A_i)\} = & \diag\big\{\varphi_{N,\sigma}[W_{N,\sigma}\varphi_{N-1,\sigma}\{W_{N-1,\sigma}\cdots\varphi_{1,\sigma}(W_{1,\sigma} A_i+b_{1\sigma})+b_{N-1,\sigma}\} + b_{N,\sigma}]\big\} \nonumber, 
\end{eqnarray}
where $b_{i,\mu}, W_{i,\mu}, b_{i,\sigma}, W_{i,\sigma} $ for $i=1,\dots, N$ are weights within the deep neural networks, $N$ is the number of layers that determine the model’s learning capacity, and $\{\varphi_{i,\mu}\}, \{\varphi_{i,\sigma}\}$ are activation functions that will be specified later. We denote these weights and activation functions together as the parameter $\phi$.  

\subsection{Derivation of the evidence lower bound in GATE}\label{subsec:apendix:elbo1}
In this part, we show how the difference between $\log p_\theta(A_i)$ and $D_{KL} (q_\phi(z_i|A_i) || p_\theta(z_i|A_i))$ can be expressed as the evidence lower bound in equation (\ref{eq:elbo:0}). 
Note that 
\begin{align*}
& D_{KL} (q_\phi(z_i|A_i) || p_\theta(z_i|A_i))  \\
 = & \int q_\phi(z_i|A_i) \log \frac{q_\phi(z_i|A_i)}{p_\theta(z_i|A_i)} d z_i = 
\int q_\phi(z_i|A_i) \log \frac{q_\phi(z_i|A_i)p_\theta (A_i)}{p_\theta(A_i|z_i) p_\theta (z_i)} d z_i \\
 = &  \log p_\theta(A_i) + D_{KL}(q_\phi(z_i|A_i) || p_\theta(z_i)) - \E_{q_\phi(z_i|A_i)} \big(\log p_\theta(A_i | z_i)\big).
\end{align*}
Then we have 
\begin{align}
& \log p_\theta(A_i) - D_{KL} (q_\phi(z_i|A_i) || p_\theta(z_i|A_i)) \nonumber \\
=& \E_{q_\phi(z_i|A_i)} [\log p_\theta(A_i | z_i)] - D_{KL}(q_\phi(z_i|A_i) || p_\theta(z_i)) :=- \mathcal{L}(A_i; \theta, \phi). \label{eq:elbo}
\end{align}

\subsection{Derivation of the evidence lower bound (ELBO) in reGATE objective function}\label{subsec:dev:elbo:regression}
In this part, we derive the ELBO from the joint log-likelihood of $(A_i,y_i)$ in equation (\ref{eq:reg_elbo}). Let $y_i$ be the trait of the $i$-th subject. The joint log likelihood of $(A_i, y_i)$ can be expressed as 
\begin{align}
\log p_\theta(A_i, y_i) = & \E_{q_\phi(z_i|A_i)} \log \frac{p_\theta (A_i, z_i)p_\theta(y_i|A_i, z_i)}{p_\theta(z_i | A_i, y_i)} \nonumber\\
 = & \E_{q_\phi(z_i|A_i)} \log \frac{p_\theta(A_i | z_i)p_\theta(z_i)p_\theta(y_i | z_i) q_\phi(z_i|A_i)}{q_\phi(z_i|A_i)p_\theta(z_i | y_i, A_i)} \nonumber\\
 = & \E_{q_\phi(z_i|A_i)}\log p_\theta(y_i|z_i) + \E_{q_\phi(z_i|A_i)} \log p_\theta(A_i | z_i)  - D_{KL}(q_\phi(z_i|A_i)||p_\theta(z_i)) \nonumber \\
 & + D_{KL}(q_\phi(z_i|A_i) || p_\theta(z_i|y_i, A_i)) \nonumber\\
  = & -  \mathcal{L}(A_i, y_i; \theta, \phi) + D_{KL}(q_\phi(z_i|A_i) || p_\theta(z_i|y_i, A_i)), \label{eq:reg_elbo} 
\end{align} 
where we have $p_\theta(y_i|A_i, z_i)=p_\theta(y_i|z_i)$ since we assume the human trait $y_i$ and the brain connectivity $A_i$ are conditionally independent given the latent representation $z_i$ for the $i$-th subject. 

\subsection{Approximation of the ELBO in reGATE objective}\label{subsec:app:elbo:regression}
Similar to Section \ref{subsec:gate_learning}, we form the Monte Carlo estimate of $\mathcal{L}(A_i, y_i; \theta, \phi)$ as 
\begin{align*}
\cL(A_i,y_i; \theta, \phi) \simeq &  \tilde{\cL}(A_i,y_i; \theta, \phi)\\ 
= &  - \frac{1}{L} \sum_{\ell=1}^L \big(\log p_\theta(A_i | z_i^{\ell}) +\log p_\theta(y_i |z_i^{\ell}) \big)+ \frac{1}{2}\sum_{k=1}^K \big(\mu_k^2 + \sigma_k^2 - 1 - \log (\sigma_k^2)\big), 
\end{align*}
and estimate $\theta, \phi$ following the stochastic variational Bayesian Algorithm \ref{table:GATE_alg} by replacing $\tilde{\cL}(A_i; \theta, \phi)$ with $\tilde{\cL}(A_i,y_i; \theta, \phi)$. 

\subsection{Derivation of the posterior distribution of $z_i$ given $y_i$}\label{subsec:post:z:y}
The posterior distribution of $z_i$ given $y_i$ can
has an explicit expression as follows: 
To achieve this goal, for each $y_i$, we need to learn the posterior distribution of $A_i \mid y_i$. In fact, the posterior computation of $A_i$ given a particular $y_i$
$p_\theta(z_i| y_i)$ and $p_\theta (A_i|z_i)$, where the latter is learned via the inference model in reGATE. 
Recall $p_\theta(z_i) \sim  N(0, I_K)$, and 
$p_\theta(y_i |z_i) \sim  N(z_i^\top \beta + b, \sigma^2)$.  
The posterior distribution of $z_i$ given $y_i$ has an explicit expression as follows: 
\begin{align*}
p_\theta(z_i | y_i ) \propto & \; p_\theta(y_i | z_i) p_\theta(z_i) \\ 
\propto &\exp\Big\{ - \frac{(y_i - z_i^\top \beta-b)^2}{2\sigma^2}-\frac{z_i^\top z_i}{2}\Big\}\\
\propto & \exp\Big\{- \frac{z_i^\top z_i + z_i^\top \beta \beta^\top z_i/\sigma^2 - 2 z_i^\top \beta (y_i-b)/\sigma^2}{2}\Big\}\\
\propto & \exp\Big\{-\frac{z_i^\top (I_K + \beta \beta^\top/\sigma^2)z_i- 2 z_i^\top \beta (y_i-b)/\sigma^2}{2}\Big\}.
\end{align*}
Therefore, we have $p_\theta (z_i | y_i) \sim N\big(\mu_z(y_i), \Sigma_z(y_i)\big)$, where 
\begin{equation}\label{eq:z_to_y}
\mu_z(y_i) = (I_K + \beta \beta^\top /\sigma^2)^{-1} \beta (y_i-b) /\sigma^2, \quad\quad \textrm{and} \quad \Sigma_z(y_i)=\big(I_K + \beta \beta^\top /\sigma^2\big)^{-1}. 
\end{equation}

\subsection{Additional real data analysis: Data visualization for HCP}
\begin{figure}[htb!]
 \centering
 \begin{tabular}{cc}  
\includegraphics[scale=0.3]{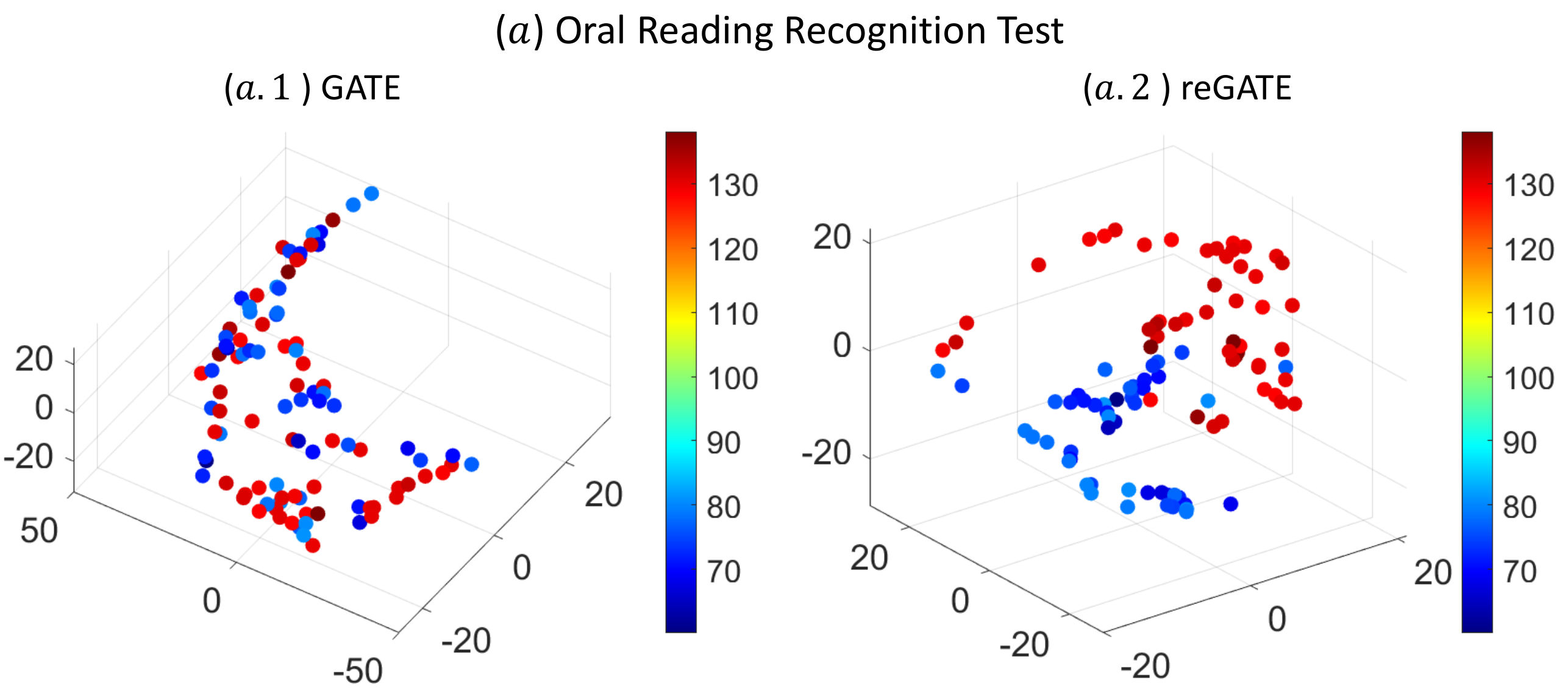}&\includegraphics[scale=0.3]{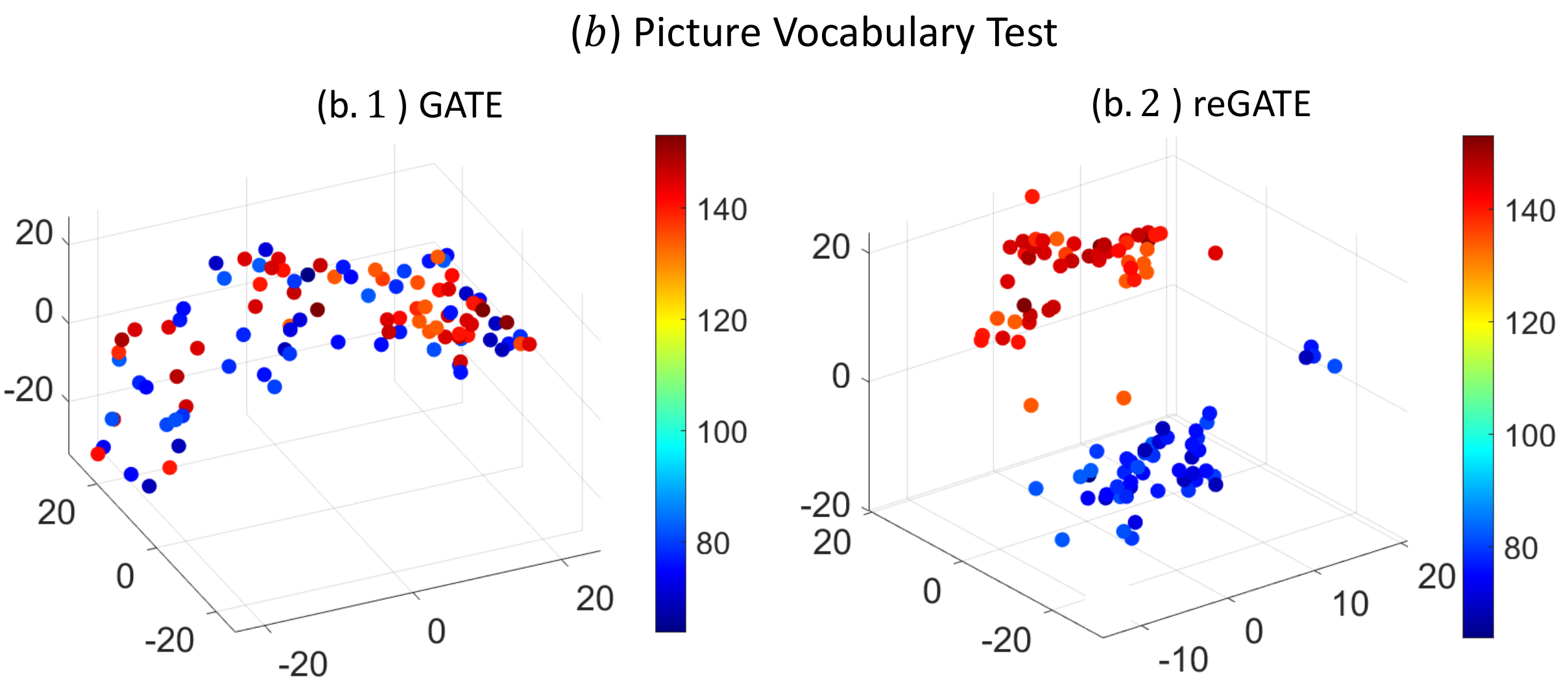} \\ 
\includegraphics[scale=0.3]{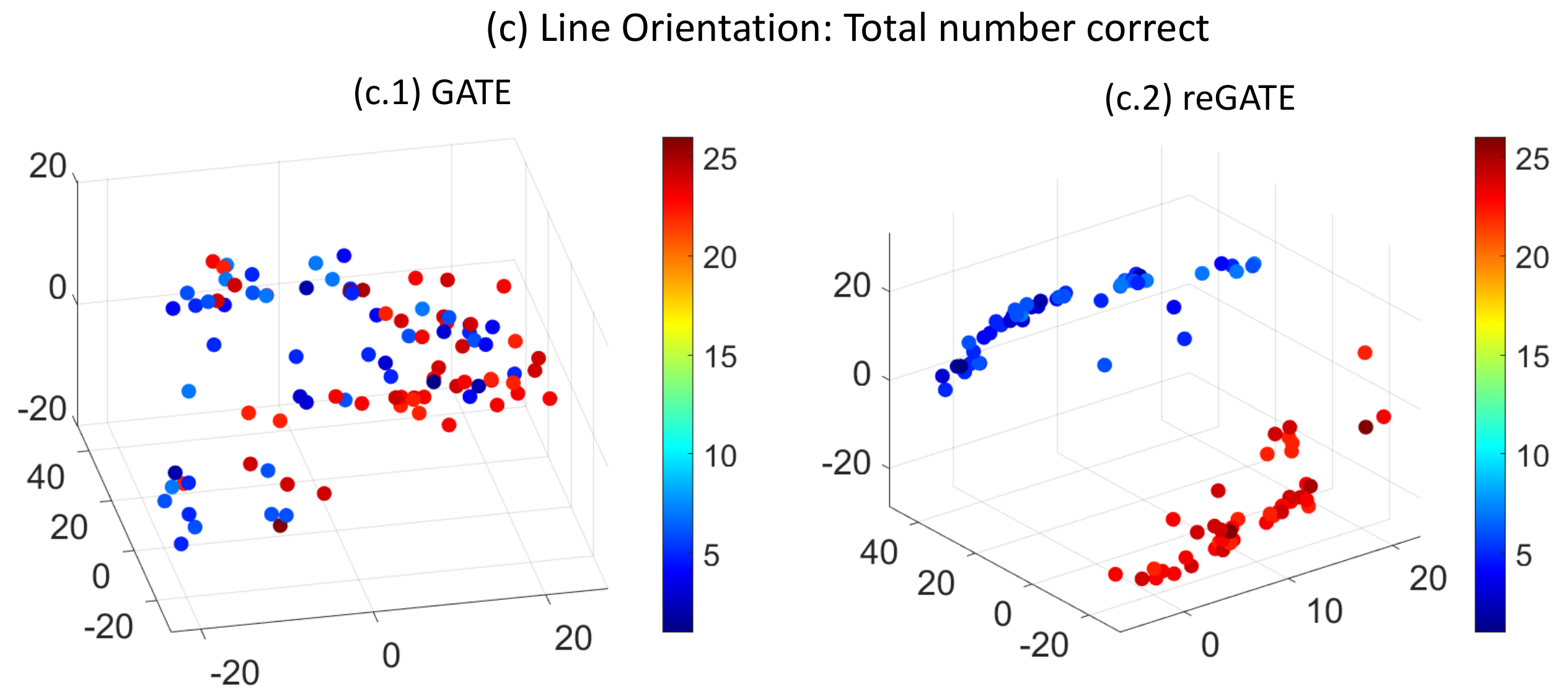}&\includegraphics[scale=0.3]{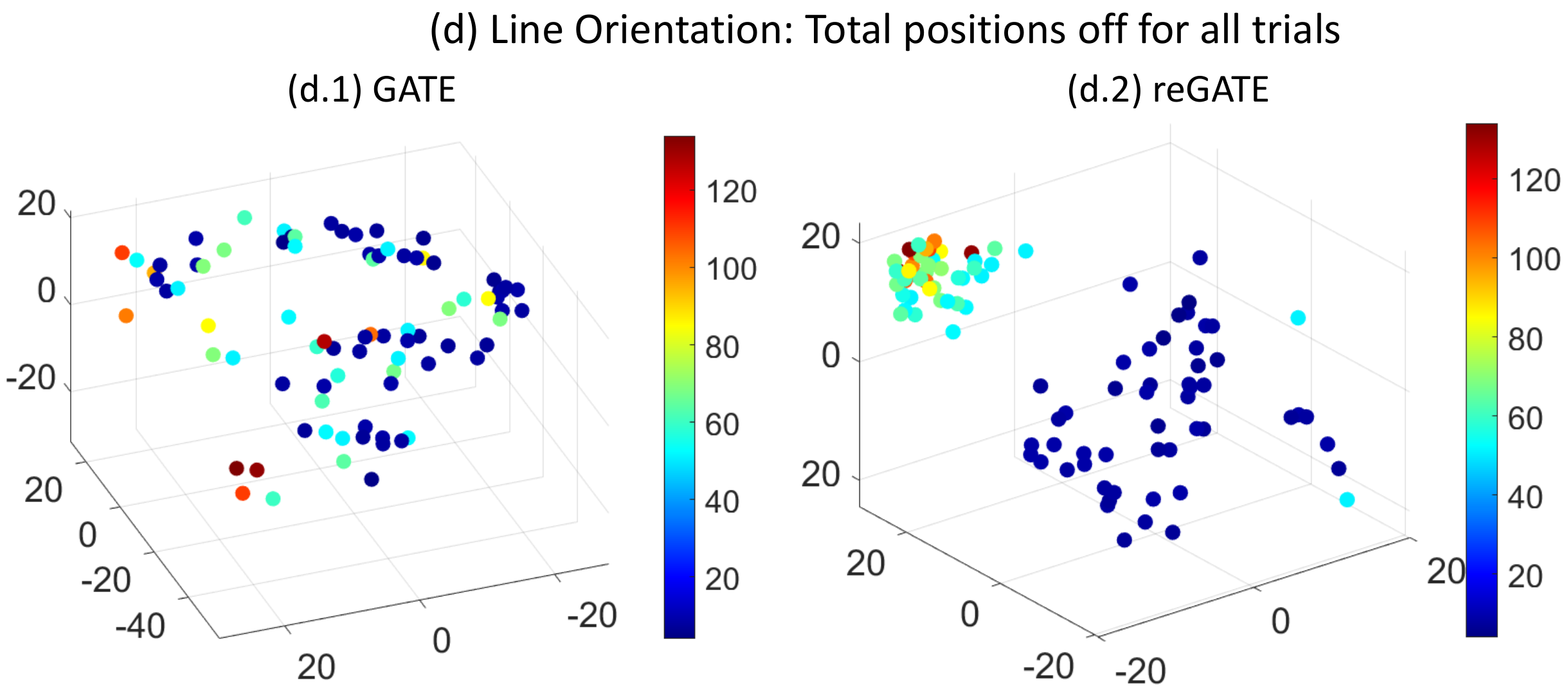} \\  
 \end{tabular}
  \caption{\small HCP study: visualization of low dimensional embedding $z_i$ learned from GATE and reGATE.  For each trait, we display 100 subjects with the lowest trait scores and 100 subjects with the highest scores. Colors represent the trait scores.
   } \label{fig:hcp:clustering}
\end{figure}

Similar to Section \ref{subsec:abcd:visua}, we first visualize the latent features of each individual's connectome in HCP, from which we can investigate the relationship between the structural connectivity and the four traits. We train GATE on $1065$ brain networks extracted from HCP to obtain low-dimensional representations $z_i$. We then plot the posterior mean of $z_i|A_i$ using t-SNE in $\mathbb{R}^3$ colored with its corresponding trait score. For each cognition trait, we only plot $200$ subjects' data, $100$ subjects with low trait scores and $100$ subjects with high scores. Figure \ref{fig:hcp:clustering} $(a)-(d)$ shows the separated map points between the two groups for different traits, and the separation is more significant under the supervised reGATE approach. 

\begin{figure}[H]
 \centering
\includegraphics[width=0.9\textwidth]{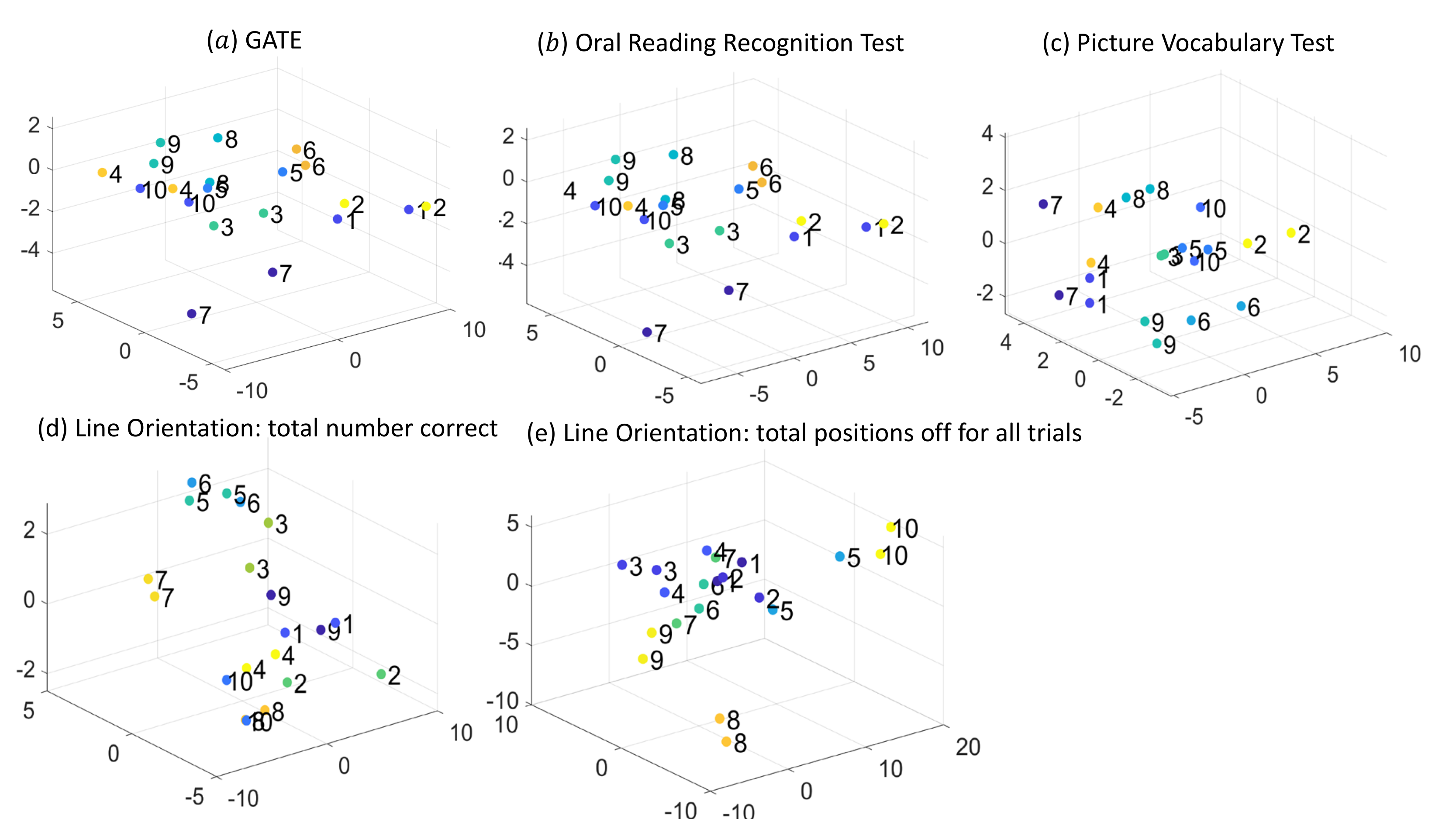}
  \caption{ Visualization of low dimensional embedding $z_i$ from the HCP test-retest data set under GATE and reGATE with four different traits. Each unique combination of color represents two scans from the same subject. 
   } \label{fig:real:test_retest}
\end{figure}

We also apply GATE and reGATE to the HCP test-retest data. Figure \ref{fig:real:test_retest} shows $10$ selected subjects, with each unique combination of color and number representing two scans from the same subject. For both  unsupervised GATE and supervised reGATE, the test-retest brain networks display a clear clustering pattern, implying that brain networks extracted from repeated scans are reproducible, and we can distinguish between different subjects based on the embedded feature $z_i$.

\subsection{Comparison between the generated networks and original data in inference}
In this section, we show the advantage of the generated network from reGATE compared with originally observed data in inference. 

\begin{figure}
 \centering
\begin{tabular}{c}
\includegraphics[height=3.5in]{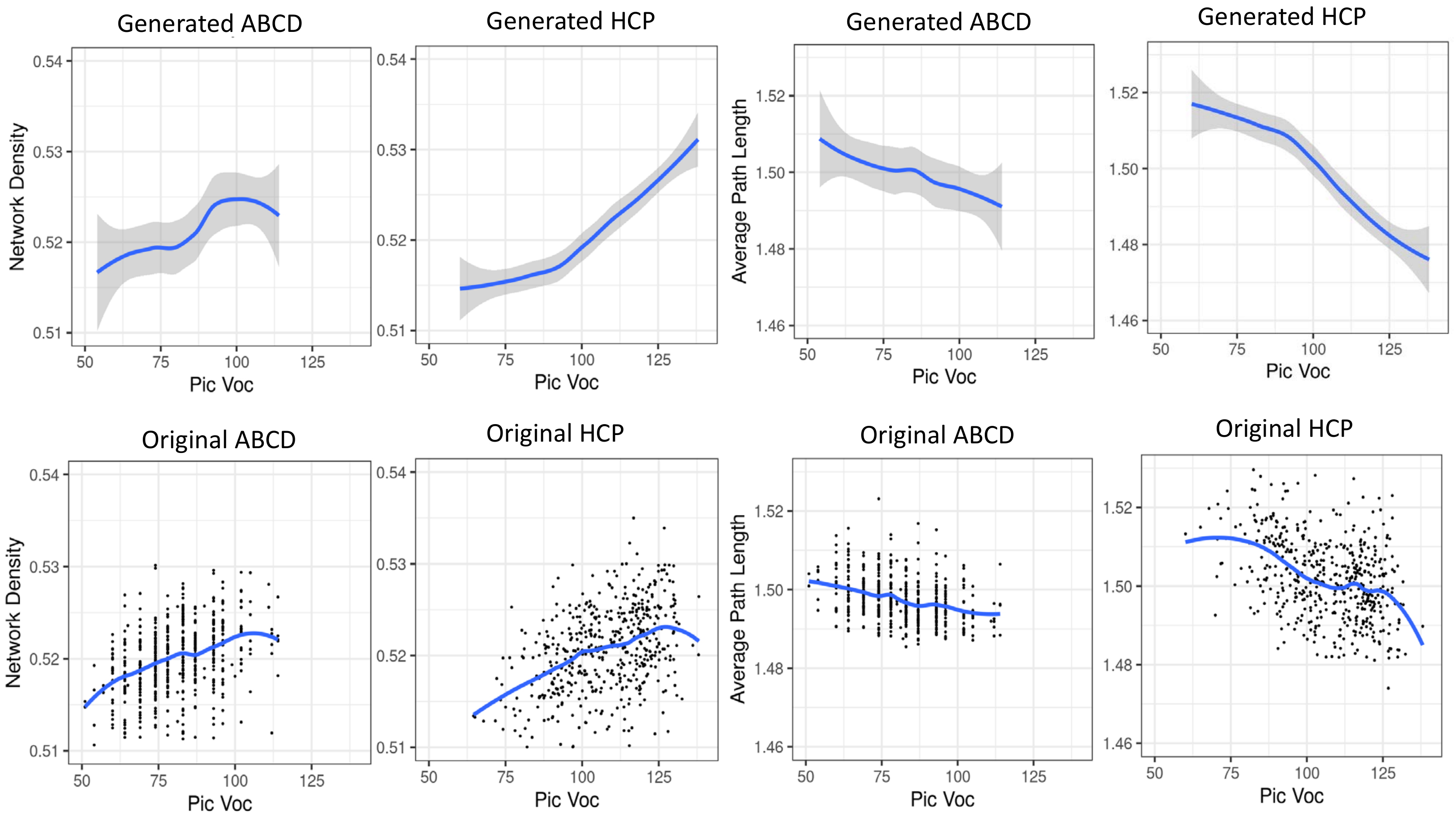}
 \end{tabular}
 \caption{ \small Comparison between generated brain networks and the observed brain networks for different $y$ in network density and average path length for ABCD and HCP datasets. 
   } \label{fig:12:compare}
 \end{figure}

It is challenging to conduct a similar inference on the real data (ABCD and HCP data) due to the limited observed brain networks for a specific picture vocabulary test score. Figure \ref{fig:12:compare} shows the comparison between the generated networks and the original datasets for the conditional prediction and confidence band of network summaries. As shown in Figure \ref{fig:12:compare}, the first row is the constructed confidence band of network density and average path length versus $y$ based on the generated networks via reGATE; see Section 4.3 for details. The second row in Figure \ref{fig:12:compare} shows the scatter plots of network density and average path length versus $y$ on the real datasets. The blue curves are the fitted mean trajectory. Since for each value of $y$, only a few or no points are observed, we cannot construct the confidence interval. It can be seen that the overall trends are similar: the network density is increasing as $y$ increases, while the average path length is decreasing as $y$ increases.

\begin{figure}
 \centering
\begin{tabular}{c}
\includegraphics[height=3.5in]{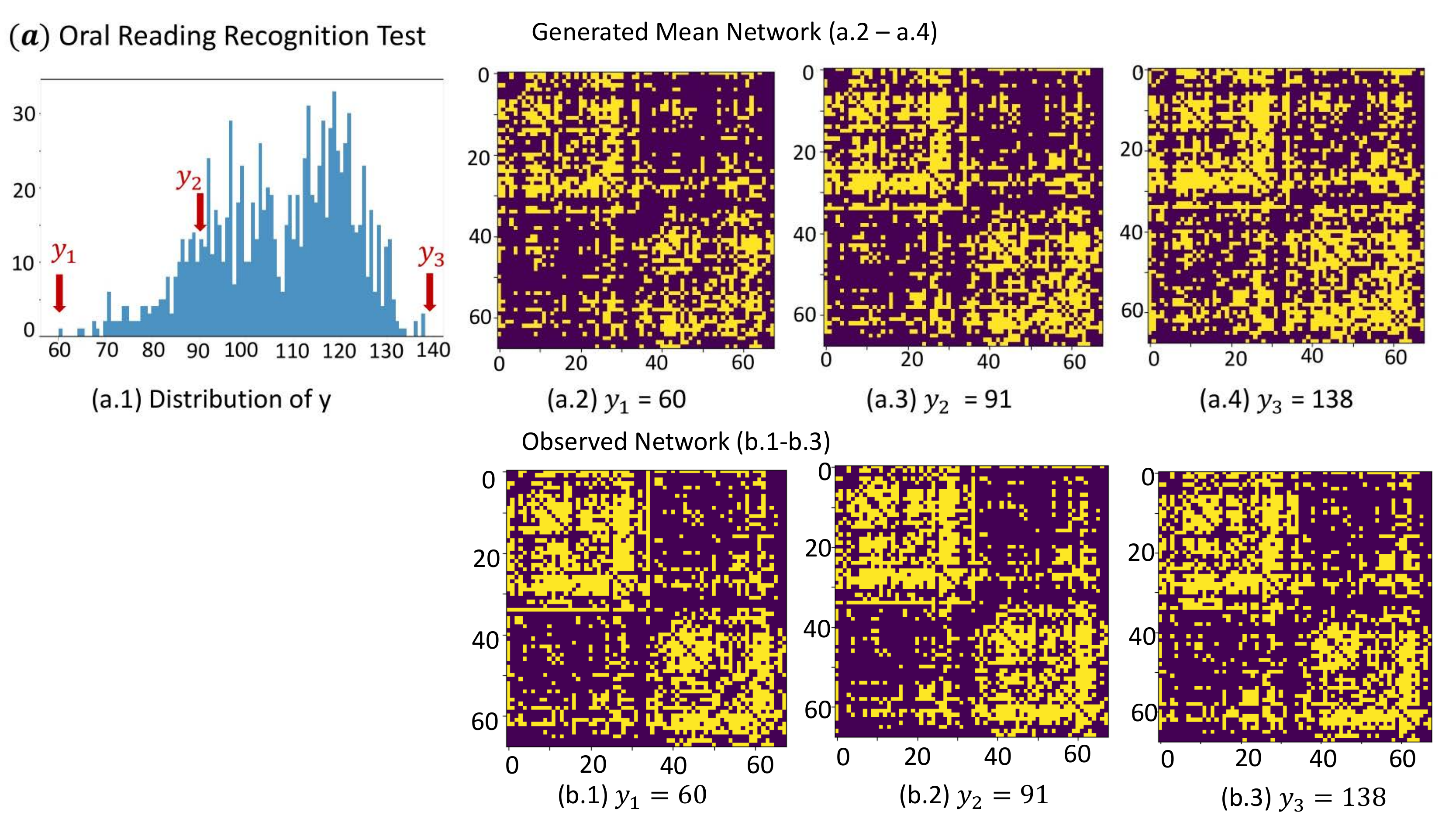}
 \end{tabular}
 \caption{ \small Comparison between generated brain networks and the observed brain networks for different $y$ in oral reading recognition test for HCP dataset. 
The first row (a.1) shows histograms of observed trait scores for the $1065$ individuals in HCP. (a.2)-(a.4) show the means of generated brain networks (after dichotomization) conditional on different $y$ under the reGATE model. (b.1)-(b.3) show the observed brain networks for different $y$ in HCP. 
   } \label{fig:13:compare}
 \end{figure}

Next, we show the comparison of Figure \ref{fig:generate:diff} with the observed dataset. We take the oral reading recognition test as $y$ in HCP as an example. We are interested to study the mean of $p(A\mid y)$ at $y=60,91$ and $138$. Figure \ref{fig:13:compare} (a.2)-(a.4) show the mean networks from a well-trained reGATE model; we refer Section 4.3 for details. From the real data, we have one subject for $y = 60$, ten subjects with  $y=91$, and two subjects for $y=138$. One can immediately see two issues in empirically estimating the posterior mean: 1) for different $y$, we have different numbers of observations; 2) we have very few data to estimate a very high-dimensional object. These two issues make the estimation less credible. Figure \ref{fig:13:compare} (b.1)-(b.3) show the mean estimated from the real data. From the results of reGATE, we can see a clear trend that increasing connectivity between the two hemispheres of the brain is correlated with better reading ability. From the result of real data, we also observe such a trend, but not so obvious compared with the reGATE's results. 

\end{document}